\documentclass{article} 
\usepackage{iclr2025_conference,times}

\usepackage{amsmath,amsfonts,bm}

\def\eqref#1{equation~\ref{#1}}

\def\1{\bm{1}}

\DeclareMathAlphabet{\mathsfit}{\encodingdefault}{\sfdefault}{m}{sl}
\SetMathAlphabet{\mathsfit}{bold}{\encodingdefault}{\sfdefault}{bx}{n}

\DeclareMathOperator*{\argmax}{arg\,max}

\usepackage{multirow}
\usepackage{hyperref}
\usepackage{url}
\usepackage{booktabs}
\usepackage{graphicx}
\usepackage{float}

\newcommand{\dataset}[1]{CAP}
\newcommand{\benchmark}[1]{PPE}

\usepackage{tcolorbox}
\tcbuselibrary{most}
\newtcolorbox{userinput}[1]{
    enhanced,
    drop shadow=black!5!white,
    left=4mm,
    right=4mm,
    top=3mm,
    bottom=3mm,
    boxsep=0mm,
    rounded corners,
    title=#1,
    fontupper=\linespread{1.1}\scriptsize\fontfamily{lmr}\selectfont
    }

\title{How to Evaluate Reward Models for RLHF}

\author{%
Evan Frick \quad Tianle Li \quad Connor Chen \quad Wei-Lin Chiang \quad Anastasios N. Angelopoulos \\
\textbf{Jiantao Jiao} \quad \textbf{Banghua Zhu} \quad \textbf{Joseph E. González} \quad \textbf{Ion Stoica} \\ \\
UC Berkeley \\
}

\iclrfinalcopy

\begin{document}

\maketitle

\begin{abstract}

We introduce a new benchmark for reward models that quantifies their ability to produce strong language models through RLHF (Reinforcement Learning from Human Feedback).
The gold-standard approach is to run a full RLHF training pipeline and directly probe downstream LLM performance.
However, this process is prohibitively expensive.
To address this, we build a predictive model of downstream LLM performance by evaluating the reward model on proxy tasks. 
These proxy tasks consist of a large-scale human preference and a verifiable correctness preference dataset, in which we measure 12 metrics across 12 domains.
To investigate which reward model metrics are most correlated to gold-standard RLHF outcomes, we launch an end-to-end RLHF experiment on a large-scale crowdsourced human preference platform to view real reward model downstream performance as ground truth. 
Ultimately, we compile our data and findings into Preference Proxy Evaluations (PPE), the first reward model benchmark explicitly linked to post-RLHF real-world human preference performance, which we open-source for public use and further development\footnote{Our code and evaluations can be found at \href{https://github.com/lmarena/PPE}{github.com/lmarena/PPE}}.


\end{abstract}

\section{Introduction}

\begin{figure}[!h]
    \centering
    \includegraphics[width=0.9\linewidth]{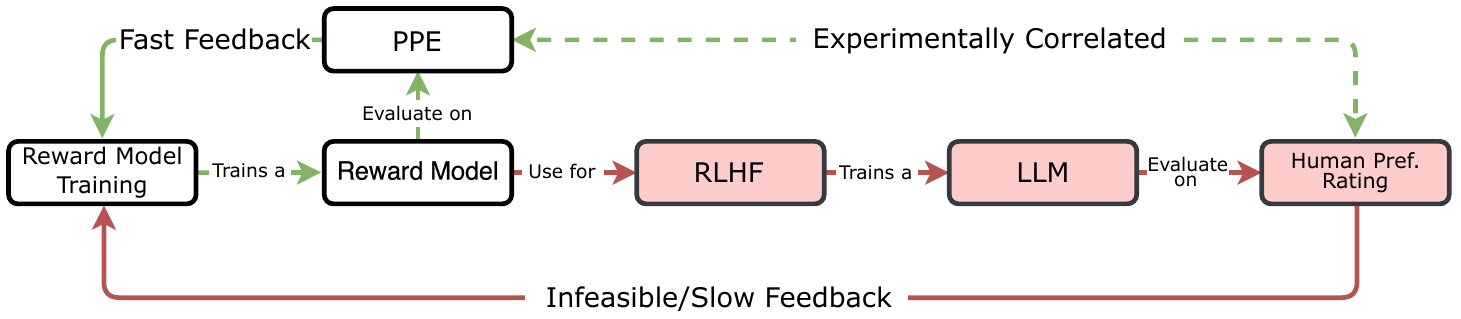}
    \caption{Overview of the RLHF pipeline. Reward models feed into the very beginning of the RLHF pipeline, making iterative improvements prohibitively slow. PPE enables a fast feedback loop that is correlated to downstream outcomes.
    }
    \label{fig:problem_overview}
\end{figure}

The ultimate test of a reward model is as follows:
\begin{center}
    Does the reward model lead to good post-RLHF language model performance?
\end{center}
In other words, because the reward model will be used as a reference signal for LLM training, in principle, only the downstream LLM performance matters.
However, to evaluate downstream performance, we must train a new LLM using the reward model and evaluate the resulting LLM---a prohibitively expensive and time-consuming process (\autoref{fig:problem_overview}). The long development-feedback cycle of reward models poses a significant challenge, limiting achievable reward model quality and, consequently, limiting the effectiveness of the entire RLHF process.

This paper introduces a cost-effective method for approximating the effect of a reward model on downstream LLM performance. Specifically, we measure reward model performance using a large-scale, crowdsourced pairwise human preference evaluation dataset, as well as a high-quality, programmatically verifiable correctness preference dataset.  To avoid introducing bias, we do not utilize LLM judges or expert annotators to provide ground-truth references. Instead, we focus on real-world preference data that reflects organic LLM usage. Additionally, we aim our evaluation tasks to closely resemble real-world RLHF training, making the assessment more aligned with practical use cases. Moreover, to bridge the existing knowledge gap between reward model evaluations and actual post-RLHF outcomes, we experimentally correlate our evaluation metrics with real human preferences on RLHF-ed LLMs. To achieve this, we used select reward models within a full RLHF training pipeline, each producing a fine-tuned LLM. These RLHF-tuned models are then deployed on Chatbot Arena where we directly measure their downstream human preference scores. Through this end-to-end analysis, we identify which metrics across diverse domains show the strongest correlation with real-world post-RLHF performance. By validating this correlation, we ensure that iterative improvements on our evaluation will lead to tangible gains in downstream performance. 

Additionally, we release \benchmark{}, a collection of 16,038 labeled human preference pairs from Chatbot Arena containing responses from 20 different top LLMs and over 121 languages as well as a dataset of 2,555 prompts, each with 32 different sampled response options, totaling 81,760 responses across 4 different models, all grounded with verifiable correctness labels.
\benchmark{} evaluates reward models on 12 different metrics and 12 different domains, such as their accuracy in selecting human-preferred or verifiably correct responses. Notably, \benchmark{} is the \textit{only} reward model benchmark directly linked to downstream RLHF outcomes.



To summarize, our work makes the following contributions:

\begin{enumerate}
    \item We analyze how reward model metrics correlate with real downstream human preference performance post-RLHF.
    \item We fully open-source \benchmark{}, a comprehensive benchmark for reward models with metrics directly linked to downstream RLHF outcomes.
\end{enumerate}

\section{Related Work}
\subsection{Human Preference and Reward Models}


Human preference has emerged as one of the gold standards for LLM training and evaluation. Several large-scale human preference datasets have been developed, including Stanford Human Preference (SHP)~\citep{pmlr-v162-ethayarajh22a}, Chatbot Arena~\citep{chiang2024chatbot}, and Anthropic HH~\citep{bai2022training}, among others. Researchers requiring human preference proxies have pursued two main approaches in this area. First, they have trained reward models based on real or synthetically generated human preference data to approximate human preferences for LLM training. Second, they have employed LLMs as judges for evaluating other LLMs.

For the training side, the line of work on Reinforcement Learning from Human Feedback (RLHF) focuses on the family of algorithms that first train a reward model as a proxy of human preferences, and then use the reward model as the signal to fine-tune the language model with reinforcement learning~\citep{christiano2023deep, bai2022training, ouyang2022training, touvron2023llama, openai2022chatgpt, bai2022constitutional, lee2023rlaif, openai2023gpt4, openai2023gpt4turbo, zhu2024starling}.

This paper studies one of the critical problems in the RLHF process: how do we evaluate reward models and select the best one for downstream performance?

\subsection{Reward Model Benchmarks}

RewardBench is the first and only previous RLHF reward model benchmark \citep{RewardBench}. RewardBench has 4 main tasks: Chat, Chat Hard, Safety, and Reasoning. The authors source considerable ground truth preference pairs from MT-Bench~\citep{zheng2023judging} and AlpacaEval~\citep{dubois2023alpacafarm}, though preference labels are also hand-verified. RewardBench also uses adversarial examples from LLMBar~\citep{zeng2024evaluatinglargelanguagemodels}, coding example pairs with correct vs buggy implementations, and safety pairs with should-refusals and should-not-refusals. Overall, RewardBench is designed to evaluate across an array of tasks posited as relevant to RLHF. RewardBench takes a crucial first step toward reward model evaluations. However, the authors assert that more research must be done to understand how to correlate performance to RLHF success. In this paper, our experiments show that as reward models have improved, we now see a negative correlation between RewardBench evaluation score on top models and downstream RLHF performance. We aim to improve upon this gap with the our findings.

\section{Sourcing Ground Truth Preference Labels}
\begin{table}[t]
\centering
\resizebox{0.9\columnwidth}{!}{
\begin{tabular}{l|l|r|r|l}
                         & Diverse Human Pref. & \multicolumn{1}{l|}{\# Prompts} & \multicolumn{1}{l|}{\# Responses} & Verified RLHF Outcomes \\ \hline
RewardBench\footnotemark & No                  & 2,985                           & 5,970                             & No                     \\
PPE (Ours)               & \textbf{Yes}        & \textbf{18,593}                 & \textbf{113,836}                  & \textbf{Yes}          
\end{tabular}}
\caption{Comparison of PPE to existing work.}
\label{tab:compare}
\end{table}
\footnotetext[1]{RewardBench is currently the only other evaluation scheme for RLHF reward models \citep{RewardBench}.}



Previous work on sourcing preference ground truth labels often relies upon LLM judge preference labels in conjunction with manual verification from individuals, introducing potential preference biases. Alternatively, rejected responses are often curated synthetically by unnaturally perturbing the chosen output or modifying the prompt to produce forced errors, introducing bias on how errors look and occur. These preference pairs are not representative of the distribution of responses seen by reward models when providing learning signals for RLHF. We offer a brief comparison to previous work in \autoref{tab:compare}.

Thus, we ground our preference labels with the following methodology: (1) Utilize crowdsourced diverse prompts and responses with human preference labels. (2) Utilize existing benchmarks with verifiable correctness checks on LLM-generated responses.

The methodology (1) provides an unbiased estimate of real-world human preference through the aggregation of many diverse human preferences. We use a large crowdsourced preference set of 16,038 preference labels to mitigate individual label noise and avoid over-fitting to any single individual's preference, details in \autoref{hp_curation}. 

Methodology (2) curates an objective correctness signal naturally unbiased by response style. We use the second approach to label the correctness of many sampled responses from an LLM, mimicking rollouts or best-of-k exploration strategies seen in RLHF training processes. As a result, we draw preference pairs from more naturally occurring distributions (eg. real LLM responses and errors), better align with the expected environment reward models operate in.

\section{Human Preference Metrics}

To measure whether a reward model aligns with human preference directly, we utilize a dataset collected from Chatbot Arena, which is a platform that allows users to vote on pairwise comparisons between responses generated from two anonymized and randomly selected LLMs. The human preference dataset contains human-labeled preferences for 16,038 pairwise comparisons between 20 selected top models\footnote{mistral-large-2402, phi-3-medium-4k-instruct, gpt-4-1106-preview, claude-3-opus-20240229, gemini-1.5-pro-api-0514, gpt-4-0314, claude-3-haiku-20240307, gpt-4-0613, claude-3-sonnet-20240229, yi-1.5-34b-chat, llama-3-8b-instruct, gemini-1.5-flash-api-0514, llama-3-70b-instruct, gpt-4o-2024-05-13, command-r-plus, gpt-4-turbo-2024-04-09, qwen2-72b-instruct, command-r, qwen1.5-72b-chat, starling-lm-7b-beta}. These models were selected based on their strong performance on Chatbot Arena and overall popularity \citep{chiang2024chatbot}. We emphasized selecting models that have already undergone some form of RLHF, anticipating that these models would be more challenging for reward models to evaluate.

Since the human preference set is crowd-sourced from Chatbot Arena, we can repeat the collection process at any time to obtain an updated set that better reflects the current array of available models and any changes in human preference. Additionally, a newly updated human preference set would largely mitigate benchmark leakage that may have occurred with the previous set. Consequently, this human preference metric can remain consistently up-to-date with fresh, relevant data.

\subsection{Curation} \label{hp_curation}

Specifically, we curate our human preference data from Chatbot Arena battles. A ``battle" consists of a user-provided prompt, two models and their responses to the prompt, and the user's preference vote for the responses. We perform a random sample weighted by model occurrence to obtain 50,000 battles from Chatbot Arena between selected models such that models are represented at a uniform frequency, then de-duplicate and remove any samples containing P.I.I information using Azure AI. 
We use OpenAI's moderation API to flag and remove potentially harmful conversations from the sample. Finally, we subsample 16,038 pairs from the remaining battles to construct the human preference dataset.

The human preference dataset, at a glance:
\begin{enumerate}
    \item Includes 4,583 instruction-following prompts, 5,195 hard prompts, 2,564 math prompts. Prompts may exist in multiple categories.
    
    \item Includes user queries from over 121 languages. Top languages include English (8,842), Chinese (1,823), Russian (1,779), German (568), Korean (338), Japanese (321), etc. 
    
    \item Includes preferences crowdsourced from 6,120 individuals.
\end{enumerate}


\subsection{Scoring}
\label{sec:hp_scoring}
We conduct several statistical metrics described below to evaluate different aspects of a given reward model. 

1. \textbf{Accuracy.} We compute pairwise ranking accuracy against a human preference label for each reward model, excluding battles in which the human rater selected a "tie". This measures the granular case-by-case similarity to a real human preference signal.

2. \textbf{Correlation.} Since each battle contains information on model identities, each reward model produces a ranking and a pairwise win-rate matrix for the 20 selected models. We compute Spearman and Kendall correlation between model ranking produced by each reward model against ground truth ranking. In addition, we compute row-wise Pearson Correlation between the win-rate matrix produced by each reward model against the ground truth win-rate matrix. We intuit that these aggregate correlation metrics measure overall similarity to real human preference.

3. \textbf{Confidence.} To weight stability in assigning preferences, we follow the metrics proposed in Arena-Hard-Auto \citep{li2024crowdsourceddatahighqualitybenchmarks}, where we measure each reward models's Separability with Confidence Interval, Confidence Agreement, and Brier Score against ground truth ranking. These metrics are designed to measure uncertainties and over-confidence within a reward model. 

Furthermore, we can calculate all the above scores conditioned on any subset of prompts in the evaluation data, specifically capturing 7 different domains. For example, we can observe these metrics on only math prompts or only instruction following prompts.  We expect that strong reward models should score high regardless of the selected domain. Scores for all subsets are detailed in Appendix \ref{appendix:A}.

\section{Correctness Metrics}

To measure a preference model's ability to distinguish between different samples drawn from the same distribution, we utilize correctness metrics on established, reputable benchmarks with verifiable ground truths (e.g. \citet{austin2021programsynthesislargelanguage}'s MBPP-Plus). We construct a dataset wherein each prompt is associated with 
32 different responses sampled from the same LLM. Additionally, since we use benchmarks with verifiable ground truths, we can score the correctness (a binary label) of each response according to the original static benchmark's 
verification function (e.g. code unit tests or Regex matching).

To assess the performance of reward models (and LLMs-as-judges), we obtain rewards/preferences for the sampled responses and evaluate how well these align with the verifiable correctness signal, with the general assumption that expert humans would always prefer correct answers over incorrect ones. Our response sampling strategy ensures that the preference labeler must disentangle the correctness signal from potentially very similar or even adversarial outputs, thereby increasing task difficulty. Moreover, this method naturally samples ``unforced" errors as they would appear in real training or evaluation schemes, rather than synthetically constructing preference pairs that may contain underlying confounding biases.



\subsection{Curation}
\label{sec:correctness_metric_curation}

For the correctness metrics, we selected standard, widely used, reputable, and verifiable benchmarks: MMLU Pro~\citep{wang2024mmlu}, MATH~\citep{hendrycksmath2021}, GPQA~\citep{rein2023gpqa}, MBPP Plus~\citep{austin2021programsynthesislargelanguage}, and IFEval~\citep{zhou2023instruction}. Each benchmark covers a different domain: general knowledge, mathematics, STEM, coding, and instruction following, respectively. While we initially curate \benchmark{} with these five benchmarks, it should be noted that any desired verifiable benchmark can be added to the correctness measurement paradigm by repeating the process outlined below, thereby providing a framework for customization towards specific evaluation needs.

For each benchmark, we sample LLM responses for 500 randomly selected prompts, each 32 times, for a total of 16,000 completions. If a benchmark has fewer than 500 prompts, we use all available prompts. We choose a large K of 32 to allow models to generate more diverse responses, covering a larger input domain for the human preference proxy and testing greater robustness to over-optimization. We note that this sampling strategy actually yields very similar KL-Divergence shifts as would be seen in RLHF training methods such as Proximal Policy Optimization (PPO) \citep{gao2022scaling, schulman2017proximal}.

We repeat this process for four different models: Llama-3-8B-Instruct, Gemma-2-9b-it, Claude-3-Haiku, and GPT-4o-mini-2024-07-18 \citep{llama3modelcard, team2024gemma, claude32024family, openai2024gpt4omini}. Each model samples prompts randomly with different seeds. We reason that different model response distributions may have different difficulties. For example, an already extremely high-performing model like GPT-4o-mini-2024-07-18 may be more challenging for reward models to evaluate correctness.

We then score all responses using the benchmark's verification methods. Using the correctness labels for all responses, we discard any rows in which the model got every single response wrong or every single response right, as it is impossible for the reward model to select a better generation in these cases. Additionally, we discard any row where less than 10\% or greater than 90\% of the responses were correct, with exceptions made for benchmarks with very few valid options. This step helps avoid vacuously correct responses, such as an LLM randomly guessing the correct multiple-choice answer with completely nonsensical reasoning, as well as prompts that are too easy.

From the remaining data, we randomly sample 128 responses from each model, totaling 512 samples. If a benchmark is too small and some models have fewer than 128 viable samples, we adjust the sampling accordingly. More details on curation can be found in Appendix \ref{appendix:smalle-bench-mod}.

\subsection{Scoring}
\label{sec:correctness_scoring}
We score the reward models on the correctness metrics in ways that target a reward model's robustness, granularity, and theoretical roof-line performance.

\subsubsection{Best of K Curves}

A best of K curve shows on average how the reward model's selected ``best" answer's ground truth score changes vs K. When plotted against the ground truth curve, we can observe the gap between the reward model's ability to select the ``best" answer given a set of K responses, and the ``gold standard" best score. More formally, let $S_K$ be a size $K$ random sample of responses from a model, $g: S_K \rightarrow \{0, 1\}$ be the ground truth scoring function, and $\hat{R}: S_K \rightarrow \mathbb{R}$ be the reward model proxy score. Then, $\mathbb{E}_{S_K}[g(\argmax_{s \in S_K}{\hat{R}(s)})]$ is the expected ground truth score of the select response by the reward model given K sampled responses. We then sweep across K = 1,..., 32 to obtain a curve.

These curves represent how much the reward model can differentiate the LLM's generations whilst picking from examples drawn from the same distribution. The simple intuition here is that as K increases, the ``exploration" of the LLM is expanded, thereby increasing the likelihood that a correct answer lies within the K different samples. However, as exploration increases, the likelihood that a response that exploits the reward model is present also increases. In all best of K metrics, we use $K=32$, providing both reasonable inference costs balanced with a significant enough exploration space to test the reward model's capabilities.


In order to distill the curves into interpretable numbers, we propose several metrics:

\begin{enumerate}

    \item \textbf{Maximum Achieved Performance:} the maximum score achieved by the reward model at any point on the best of K curve. 
    Note that the maximum achieved performance is relatively agnostic to over-optimization.

    \item \textbf{Error With Respect to Ground Truth:} the expected squared error between the score of the reward model's selected response against the ground truth best response. Once again, let $S_K$ be a size $K$ random sample of responses from a model, $g: S_K \rightarrow \{0, 1\}$ be the ground truth scoring function, and $\hat{R}: S_K \rightarrow \mathbb{R}$ be the reward model proxy score. Then, the error with respect to ground truth is $\frac{1}{32}\sum_{K = 1}^{32}\mathbb{E}_{S_K}[(g(\argmax_{s \in S_K}{\hat{R}(s)}) - \max_{s \in S_K}{g(s)})^2]$

    \item \textbf{End Score:} We also look at the final score achieved by the reward model at $K=32$. If no over-fitting has occurred this should also be the maximum achieved performance.
\end{enumerate}
Best of K scores are detailed in Appendix \autoref{tab:rm_bok_performance}.

\subsubsection{Area Under Receiver Operator Characteristics (ROC) Curve}

Since the ground truth verification outputs a binary label, we can check each reward model's strength as a binary correctness classifier by calculating the area under the ROC curve. We first normalize the scores in each row with min-max normalization. Then we calculate the binary classification ROC curve using the normalized scores as ``probabilities". AUC scores are detailed in Appendix \autoref{tab:area_roc_curve}.

\subsubsection{Accuracy}

Since LLM-as-a-judge cannot easily scale 32-wise judgments, we create a supplemental pairwise task to evaluate correctness preference accuracy compatible with both reward models and LLM-as-a-judge.  For each row of best of K data, we simply sample 5 pairs of responses such that in each pair, there is one correct response and one incorrect response. Then, after randomizing positions, the LLM-as-a-judge picks the preferred response. We then measure the accuracy as the rate in which the correct response is preferred over the incorrect result. The accuracies for reward models are also collected for comparison. All scores are documented in Appendix \autoref{tab:supplemental_acc}.


\begin{figure}[t]
    \centering
    \includegraphics[width=0.9\linewidth]{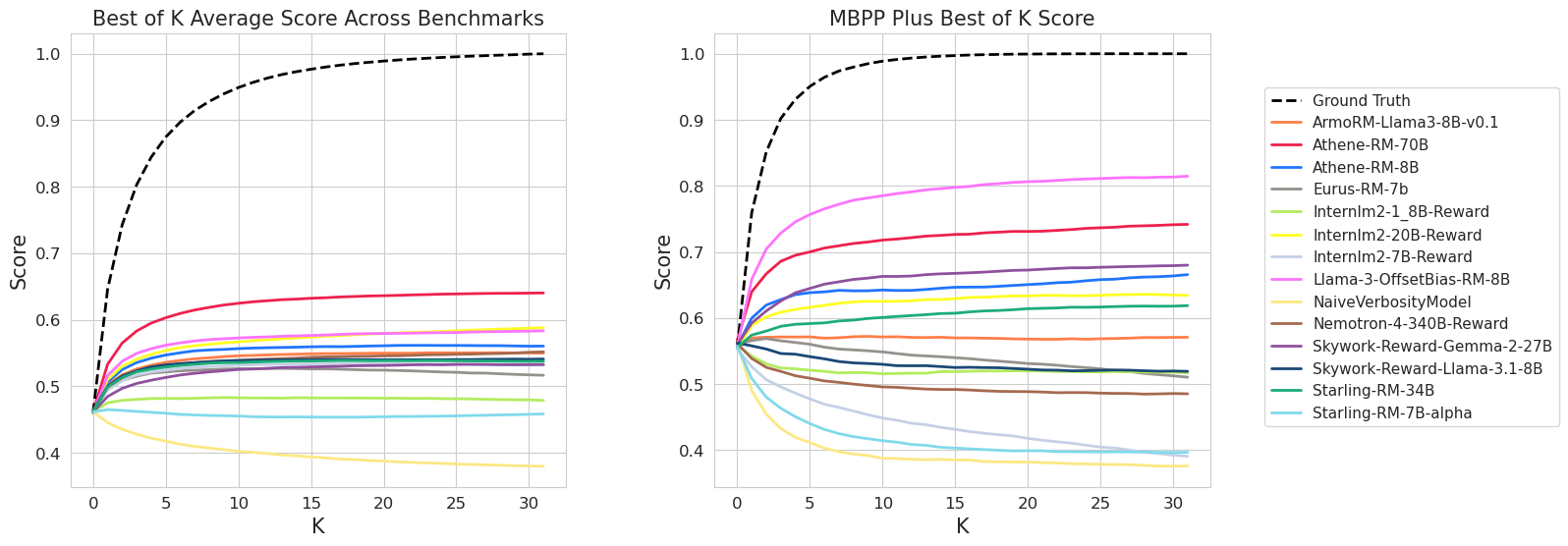}
    \caption{Example best of K curves. Left: Average performance across all benchmarks. Right: Performance on MBPP Plus benchmark showing many reward models over-optimizing. Both graphs show model average picked score vs. number of samples (K).}
    \label{fig:bok}
\end{figure}

\begin{table}[t]
    \centering
    \resizebox{0.75\columnwidth}{!}{
    \begin{tabular}{lcccccc}
    \toprule
    Reward Model & MMLU-Pro & MATH & GPQA & MBPP-Plus & IFEval & Mean \\
    \midrule
    Athene-RM-70B & 0.77 & 0.79 & 0.59 & 0.68 & 0.62 & \textbf{0.69} \\
    Claude 3.5 (ArenaHard)\textsuperscript{\textdagger} & \textbf{0.81} & \textbf{0.86} & \textbf{0.63} & 0.54 & 0.58 & 0.68 \\
    Llama-3-OffsetBias-RM-8B & 0.62 & 0.68 & 0.55 & \textbf{0.74} & 0.62 & 0.64 \\
    GPT-4o-mini (ArenaHard)\textsuperscript{\textdagger} & 0.71 & 0.81 & 0.57 & 0.54 & 0.56 & 0.63 \\
    Llama-3.1-70B (ArenaHard)\textsuperscript{\textdagger} & 0.73 & 0.73 & 0.56 & 0.58 & 0.56 & 0.63 \\
    internLM2-20B-Reward & 0.68 & 0.70 & 0.57 & 0.58 & 0.62 & 0.63 \\
    Athene-RM-8B & 0.68 & 0.71 & 0.55 & 0.62 & 0.57 & 0.62 \\
    ArmoRM-Llama3-8B-v0.1 & 0.66 & 0.71 & 0.57 & 0.54 & 0.58 & 0.61 \\
    Skywork-Reward-Llama-3.1-8B & 0.64 & 0.70 & 0.57 & 0.52 & 0.61 & 0.61 \\
    Nemotron-4-340B-Reward & 0.70 & 0.65 & 0.57 & 0.49 & 0.63 & 0.61 \\
    internLM2-7B-Reward & 0.67 & 0.73 & 0.55 & 0.44 & \textbf{0.64} & 0.60 \\
    Llama-3.1-70B (Alpaca)\textsuperscript{\textdagger} & 0.66 & 0.66 & 0.56 & 0.52 & 0.56 & 0.59 \\
    Claude 3.5 (Alpaca)\textsuperscript{\textdagger} & 0.66 & 0.63 & 0.56 & 0.52 & 0.57 & 0.59 \\
    Skywork-Reward-Gemma-2-27B & 0.54 & 0.63 & 0.53 & 0.59 & 0.54 & 0.56 \\
    GPT-4o-mini (Alpaca)\textsuperscript{\textdagger} & 0.57 & 0.64 & 0.53 & 0.52 & 0.56 & 0.56 \\
    NaiveVerbosityModel & 0.48 & 0.50 & 0.48 & 0.31 & 0.52 & 0.46 \\
    \bottomrule
    \end{tabular}}
    \caption{Reward model and LLM-as-a-judge scores on the correctness accuracy metric. LLM-as-a-judge is marked with \textdagger.}
    \label{tab:supplemental_acc}
\end{table}



\section{Validating \benchmark{} on Post-RLHF Outcomes}
\label{sec:experiment_setup}
By testing a reward model performance on a benchmark, we hope to glean insight towards downstream performance on an LLM RLHF-ed using a given reward model. To measure how well different metrics in \benchmark{} correlate to post-RLHF LLM performance on real-world human preference, we conduct an experiment in which we RLHF a given base LLM using different reward models. We then measure the real-world human preference scores of the resulting LLMs to understand the true performance of the original reward models.

For our experimental setup, we use the reward models to RLHF Llama-3.1-8B-Instruct using Direct Preference Optimization (DPO) \citep{rafailov2023direct}. This way, we can compare LLMs tuned on identical RLHF pipelines, except for the reward model being measured. Then, these RLHF-ed LLMs are deployed to Chatbot Arena to collect real-world human pairwise preferences between model answers.  Overall, 12,190 human votes were collected and compiled into relative rankings between these RLHF-ed LLMs. Under this controlled RLHF experiment, the non-noise variance in final human preference rankings attained by these models is dependent only on the reward model choice, effectively measuring the downstream performance of these reward models, albeit on a single model base model undergoing off-policy DPO RL training.

\subsection{Training Procedure}

Nine\footnote{Selected: Athene-RM-70B and Athene-RM-8B, InternLM2-20B-Reward, InternLM2-7B-Reward, Llama-3-OffsetBias-RM-8B, ArmoRM-Llama3-8B-v0.1, Skywork-Reward-Gemma-2-27B, Skywork-Reward-Llama-3.1-8B, Nemotron-4-340B-Reward \citep{athene70b2024, cai2024internlm2, park2024offsetbias, ArmoRM, skyworkreward2024, wang2024helpsteer2}.  Evaluated on 
Preference Proxy Evaluations (PPE), but not selected: Starling-RM-34B, Starling-RM-7B-Alpha, Eurus-RM-7B, InternLM2-1.8B-Reward, and NaiveVerbosityModel \citep{starling2023, yuan2024advancing, cai2024internlm2}.} reward models were selected to act as preference labels in a full RLHF training pipeline in which the resulting models were evaluated on real human preference. We constrained this experiment to nine models for cost reasons-- the RLHF and human preference evaluation process is exceedingly expensive. We selected popular, newer, and high-performing reward models from RewardBench. We reason these will be the most difficult reward models to differentiate. We also require the selected reward models to be general-purpose reward models, and not specifically tuned to any single domain or task.

We create a training dataset by first including 7,000 prompts sampled from the original 50,000 human preference votes after PII removal, unsafe prompt removal, and de-duplication. We then add 500 random prompts from MMLU-Pro that are not in \benchmark{}, and another 500 prompts from MATH train set (also mutually exclusive from \benchmark{}). For each prompt, we sample 16 responses from the base model, Llama-3.1-8B-Instruct, randomizing the temperature for each generation, drawing from a triangular distribution ($a=0.0, b=1.0, c=1.3$) to promote more diverse exploration. This process yields 8,000 total prompts, each with 16 different responses, totaling 128,000 responses.

Each reward model then constructs its own preference dataset. First, the reward model gives scores for each of the 16 responses for each prompt. The ``chosen" response is set as the maximum scoring response. The ``rejected" response is sampled as the rank $n$ response, where $n$ is sampled uniformly. Note that the sample for $n$ is seeded such that it is the same for each across reward models. This process yields a dataset of 8,000 rows, each with a prompt, a chosen response, and a rejected response where both responses are in-distribution for the base model-- a requirement for using DPO.

We then train Llama-3.1-8B-Instruct on each dataset using DPO producing an LLM associated with each selected reward model for real-world downstream human preference testing.

\subsection{Evaluation on Real-World Human Preference}\label{arena_experiment}

\begin{table}[t]
\centering
\resizebox{0.65\columnwidth}{!}{
\begin{tabular}{lrrr}
\toprule
Model & Arena Score & 95\% CI Lower & 95\% CI Upper \\
\midrule
Meta-Llama-3.1-70B-Instruct\textsuperscript{*} & 1228 & 1218 & 1238 \\
Athene-RM-70B & \textbf{1216} & 1206 & 1226 \\
Athene-RM-8B & 1209 & 1199 & 1219 \\
InternLM2-7B-Reward & 1204 & 1194 & 1212 \\
Llama-3-OffsetBias-RM-8B & 1200 & 1191 & 1209 \\
ArmoRM-Llama3-8B-v0.1 & 1189 & 1181 & 1198 \\
Meta-Llama-3.1-8B-Instruct\textsuperscript{*} & 1178 & 1168 & 1187 \\
Skywork-Reward-Llama-3.1-8B & 1176 & 1166 & 1185 \\
Skywork-Reward-Gemma-2-27B & 1173 & 1163 & 1182 \\
InternLM2-20B-Reward & 1173 & 1163 & 1182 \\
Nemotron-4-340B-Reward & 1172 & 1163 & 1180 \\
Meta-Llama-3-8B-Instruct\textsuperscript{*} & 1152 & 1143 & 1162 \\
\bottomrule
\end{tabular}}
\caption{Post DPO performance on Chatbot Arena Overall Category. ``Model" is the reward model used to train the base model. Models marked with ``*" are baseline unaltered models. The best non-base model Arena Score is bolded.}
\vspace{-0.5em}
\label{tab:chatbot_area_elos_overall}
\end{table}

We deploy the trained models to Chatbot Arena to undergo blind evaluation from real users. We set up a cohort of 13 models which include the trained DPO models as well as Llama-3.1-8B-Instruct, Llama-3.1-70b-Instruct, and Llama-3-8B-Instruct. All models used temperature $0.2$ (excluding Llama-3-8B-Instruct at temperature 0.7). Model pairs were sampled evenly with only each other for battles. Battles were collected over a six day period, from September 10th, 2024 to September 16th, 2024.  In all battles, the receiving user was selected randomly. Additionally, the model names (labeled \texttt{llama-3.1-8b-dpo-test-\{1,2...,9\}}) were not revealed to the user until after the vote was given.

Overall, 12,190 human preference votes were collected, with an average of 2,032 battles per model, and an average of 190 battles per unique model pair. More details on battle statistics and be found in Appendix \autoref{tab:human-pref-votes} of Appendix \ref{appendix: C}. The resulting Arena Scores are detailed in \autoref{tab:chatbot_area_elos_overall}. Arena Scores are calculated using the Bradley-Terry model, as proposed in \citet{chiang2024chatbot}.


\section{Studying Correlation with Downstream Performance} \label{downstream-analysis}
\begin{figure}[t]
    \centering
    \includegraphics[width=.9\linewidth]{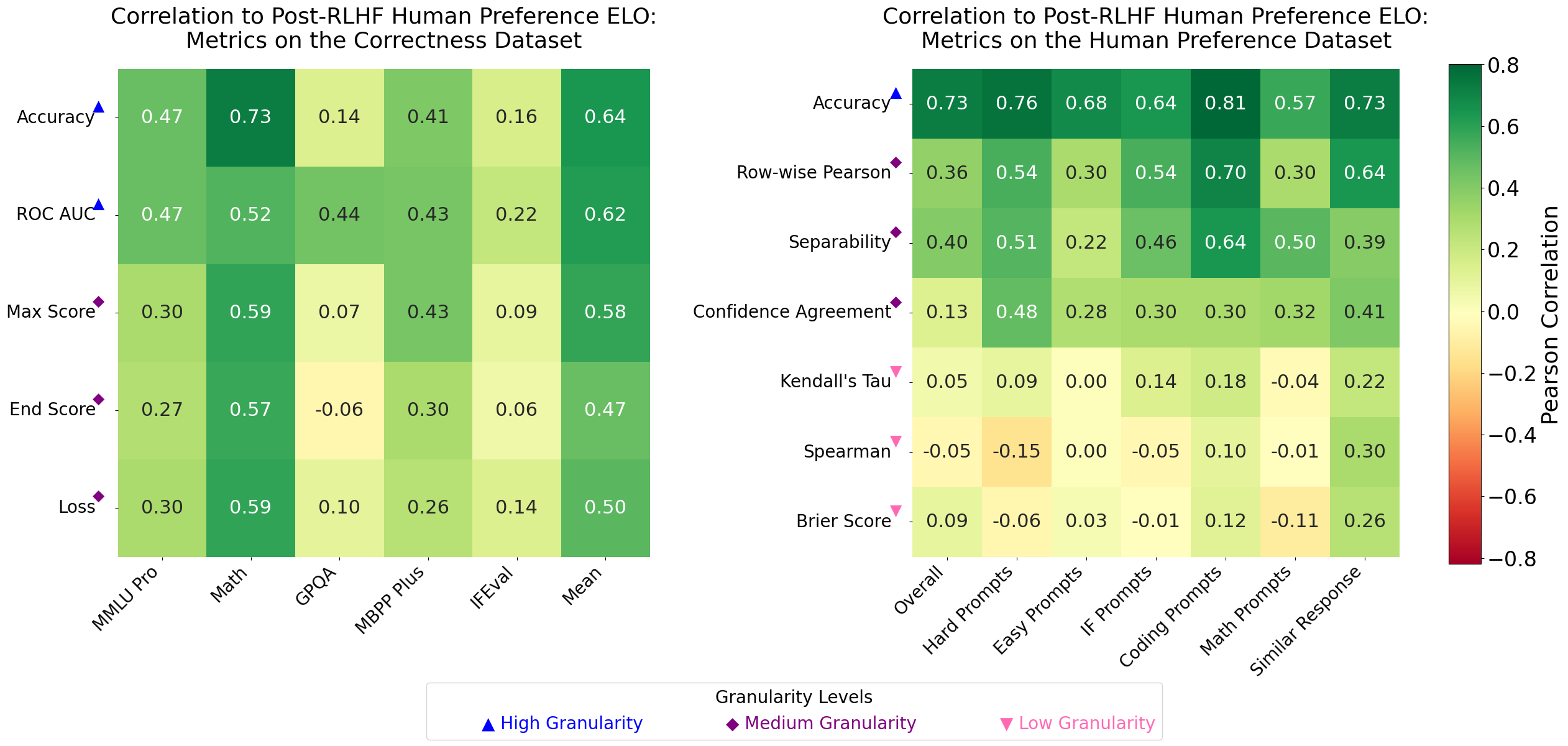}
    \caption{Pearson correlations of different metrics toward downstream human preference scores. Left: Pearson correlation between the ranking of models on 5 specific benchmarks and 5 different metrics and their respective post-DPO Arena Score rankings on real human preference. Right: Pearson correlation between the ranking of models on 7 categories and 7 metrics on the Human Preference Dataset. A similar version using style controlled human preference Arena Scores as reference is shown in Appendix \autoref{fig:sccorrs-cap}.}
    \label{fig:corrs-cap}
\end{figure}

\begin{figure}[t]
    \centering
    \includegraphics[width=0.4\linewidth]{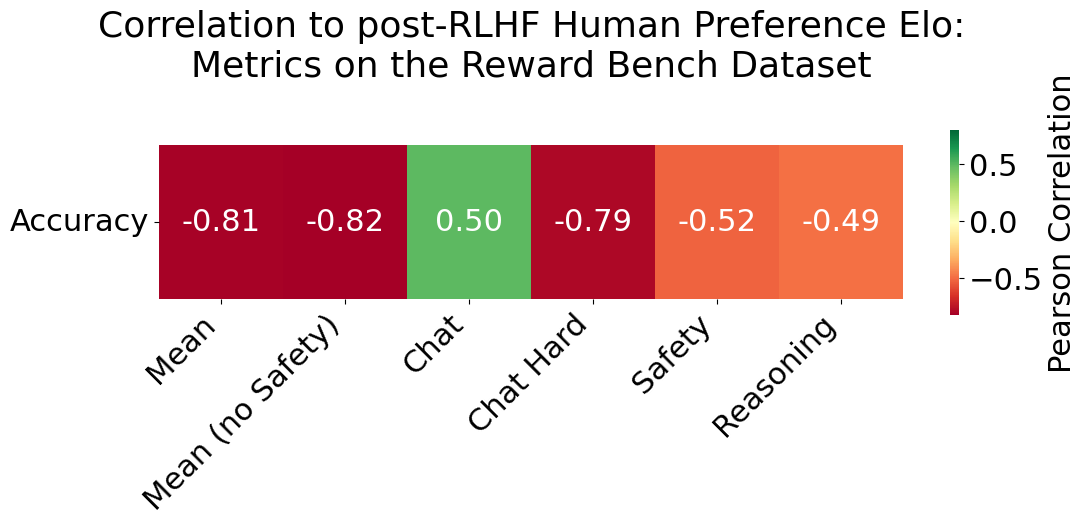}
    \caption{Pearson correlation between the ranking of models in RewardBench and their
    respective post-DPO Arena Score rankings on real human preference. Style controlled version in Appendix \autoref{fig:screward-bench-correlations}.}
    \label{fig:reward-bench-correlations}
\end{figure}

\begin{figure}[t]
    \centering
    \includegraphics[width=.80\linewidth]{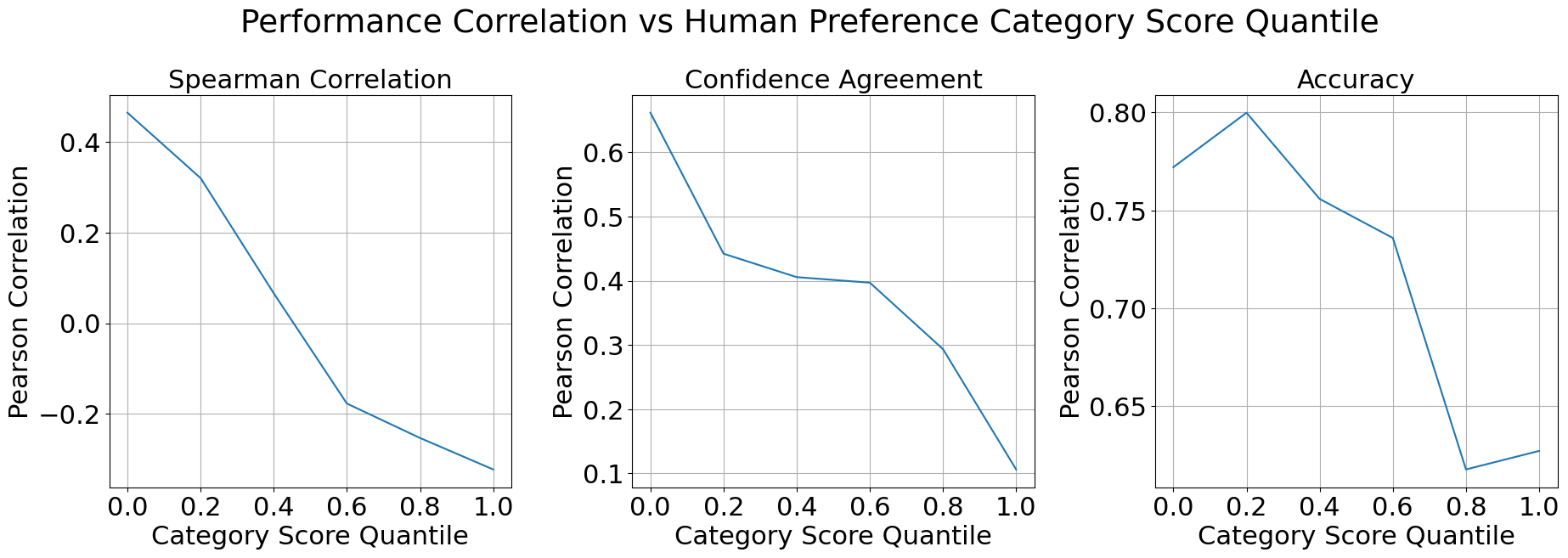}
    \caption{Spearman Correlation, Confidence Agreement, and Accuracy metrics:
For each metric, we take the quantiles of category scores (Hard, Easy, Instruction Following, Coding, Math, and Similar). The Pearson Correlation is calculated relative to Post-RLHF Human Preference Arena Score ratings for each quantile. Notably, accuracy peaks at 0.80 correlation at low quantile aggregation.}
    \label{fig:quantiles}
\end{figure}

In this section, we analyze how different metrics correlate with post-RLHF human preference scores (experimental setup detailed in Section \ref{arena_experiment}).
Our main results are displayed in Figure~\ref{fig:corrs-cap}, which shows the correlations of our offline reward model evaluations against the real-world human-preference ranking from the crowdsourced platform.

On correctness metrics (left plot in \autoref{fig:corrs-cap}) we make several observations: (1) Mean across all domains is well correlated across all metrics, but exhibits higher correlation with AUC and Accuracy scores. (2) Math is the best individual benchmark domain in terms of predictive power. (3) ROC AUC score draws higher correlation across all benchmarks, even on benchmarks that are otherwise uncorrelated.

Turning to the right-hand side of Figure~\ref{fig:corrs-cap}, the accuracy of the reward model is the best predictor of the fine-tuned LLM's preference score. Row-wise Pearson Correlation, Confidence Agreement, and Separability show some correlative power to downstream Arena Score but do not exceed accuracy. Meanwhile, metrics like the Spearman correlation and Kendall correlation have nearly zero correlation with the final Arena Score achieved by the post-DPO models. One possible reason for this trend is that accuracy measures expected preference correctness per preference pair--- a much more granular scale. Other metrics involve aggregating reward model signals over higher-order preferences, such as preference for each model, as measured by correlation metrics. We consider these metrics as low granularity. Medium granularity metrics, such as Row-wise Pearson Correlation aggregate reward model signal, but do so over smaller subsets of preferences.



Overall, accuracy on the human preference dataset is more correlated than the correctness metrics. This is because correctness and human preference do not necessarily align. Moreover, the information contained in Loss, Max score, and End score may not prove relevant in DPO, which is off-policy. Those employing RLHF algorithms that have a higher risk of over-optimization may find these alternative measures helpful. However, when calculating correlation against style controlled Arena Scores\footnote{Style controlled Arena Scores are calculated as detailed in \citet{stylearena2024}.} we notice a slight decrease in correlations on the human preference dataset. Notably, the correctness preference measurements show no change, suggesting correctness preference may be more robust towards reward model preference quality, response style aside. We leave details for Appendix \ref{appd-style-control}.


Additionally, we observe that measuring the lower bound score may correlate more to downstream RLHF performance than the average score or upper bound score. In \autoref{fig:quantiles}, we first re-scale each category's scores to be mean $0$ and SD $1$, then we vary the quantile of the aggregation strategy across human preference dataset categories seen in Appendix \autoref{tab:overall_cap_table} (Hard Prompts, Easy Prompts, etc). In this case, the $0$ quantile is the minimum, and the $1$ quantile is the maximum. We find that in nearly every metric, decreasing the quantile increases correlation with downstream Arena Scores. We posit this represents the requirement that reward models be robust under all input distributions. Any domain weakness in a reward model can be exploited by the LLM during training.

\section{Limitations}

\subsection{Benchmark Leakage}

We acknowledge that benchmark leakage is a very real possibility. We also consider two factors that help mitigate this issue: (1) The human preference dataset can be updated with new crowdsourced preference data at any time. This includes adapting to the most recent prompt and response distributions. (2) The correctness preference datasets can be extended to any other benchmark that becomes standard enough to be widely used.


\subsection{Limits on Testing Downstream Performance}

Unfortunately, end-to-end evaluation of reward models via post-RLHF LLM performance on human preference is extremely expensive and time-consuming. As such, we are limited to testing the performance of nine select models, rather than all reward models. In addition, we use DPO, an offline RL algorithm over PPO, an online algorithm, which may play more into over-optimization issues or may have different reward model requirements altogether. Therefore, we note that the downstream performance is in the context of the base model and RLHF learning algorithm used, and is not a unilateral measurement of downstream outcomes in all possible cases. Future work should experimentally verify the desired reward model behavior of other RLHF methods.



\section{Conclusion}


We present PPE, a reward model benchmark explicitly tied to post-RLHF outcomes based on real human preferences. Our experiment aims to identify which metrics, applied to specific tasks, correlate most strongly with downstream performance. We find that across the board, granular measurements, such as accuracy, are the best predictors. Additionally, our results suggest that measuring lower bound performance may be more indicative of expected reward model performance in the RLHF pipeline. Overall, our evaluations achieve a 77\% Pearson correlation with downstream performance, significantly improving upon previous work. Based on these results, we encourage future research to further investigate reward model quality and downstream RLHF performance under broader conditions. We fully open-source dataset creation, experimental validation, and reward model evaluation code and methods. We anticipate that the high-quality preference evaluation in PPE, combined with our post-RLHF analysis of metric predictive power, will significantly advance vital research into reward models and RLHF.

\newpage
\bibliography{iclr2025_conference}

\begin{thebibliography}{40}
\providecommand{\natexlab}[1]{#1}
\providecommand{\url}[1]{\texttt{#1}}
\expandafter\ifx\csname urlstyle\endcsname\relax
  \providecommand{\doi}[1]{doi: #1}\else
  \providecommand{\doi}{doi: \begingroup \urlstyle{rm}\Url}\fi

\bibitem[AI@Meta(2024)]{llama3modelcard}
AI@Meta.
\newblock Llama 3 model card.
\newblock 2024.
\newblock URL \url{https://github.com/meta-llama/llama3/blob/main/MODEL_CARD.md}.

\bibitem[Anthropic(2024)]{claude32024family}
Anthropic.
\newblock The claude 3 model family: Opus, sonnet, haiku.
\newblock \url{https://www-cdn.anthropic.com/de8ba9b01c9ab7cbabf5c33b80b7bbc618857627/Model_Card_Claude_3.pdf}, 2024.
\newblock (Accessed on 06/05/2024).

\bibitem[Austin et~al.(2021)Austin, Odena, Nye, Bosma, Michalewski, Dohan, Jiang, Cai, Terry, Le, and Sutton]{austin2021programsynthesislargelanguage}
Jacob Austin, Augustus Odena, Maxwell Nye, Maarten Bosma, Henryk Michalewski, David Dohan, Ellen Jiang, Carrie Cai, Michael Terry, Quoc Le, and Charles Sutton.
\newblock Program synthesis with large language models, 2021.
\newblock URL \url{https://arxiv.org/abs/2108.07732}.

\bibitem[Bai et~al.(2022{\natexlab{a}})Bai, Jones, Ndousse, Askell, Chen, DasSarma, Drain, Fort, Ganguli, Henighan, et~al.]{bai2022training}
Yuntao Bai, Andy Jones, Kamal Ndousse, Amanda Askell, Anna Chen, Nova DasSarma, Dawn Drain, Stanislav Fort, Deep Ganguli, Tom Henighan, et~al.
\newblock Training a helpful and harmless assistant with reinforcement learning from human feedback.
\newblock \emph{arXiv preprint arXiv:2204.05862}, 2022{\natexlab{a}}.

\bibitem[Bai et~al.(2022{\natexlab{b}})Bai, Kadavath, Kundu, Askell, Kernion, Jones, Chen, Goldie, Mirhoseini, McKinnon, Chen, Olsson, Olah, Hernandez, Drain, Ganguli, Li, Tran-Johnson, Perez, Kerr, Mueller, Ladish, Landau, Ndousse, Lukosuite, Lovitt, Sellitto, Elhage, Schiefer, Mercado, DasSarma, Lasenby, Larson, Ringer, Johnston, Kravec, Showk, Fort, Lanham, Telleen-Lawton, Conerly, Henighan, Hume, Bowman, Hatfield-Dodds, Mann, Amodei, Joseph, McCandlish, Brown, and Kaplan]{bai2022constitutional}
Yuntao Bai, Saurav Kadavath, Sandipan Kundu, Amanda Askell, Jackson Kernion, Andy Jones, Anna Chen, Anna Goldie, Azalia Mirhoseini, Cameron McKinnon, Carol Chen, Catherine Olsson, Christopher Olah, Danny Hernandez, Dawn Drain, Deep Ganguli, Dustin Li, Eli Tran-Johnson, Ethan Perez, Jamie Kerr, Jared Mueller, Jeffrey Ladish, Joshua Landau, Kamal Ndousse, Kamile Lukosuite, Liane Lovitt, Michael Sellitto, Nelson Elhage, Nicholas Schiefer, Noemi Mercado, Nova DasSarma, Robert Lasenby, Robin Larson, Sam Ringer, Scott Johnston, Shauna Kravec, Sheer~El Showk, Stanislav Fort, Tamera Lanham, Timothy Telleen-Lawton, Tom Conerly, Tom Henighan, Tristan Hume, Samuel~R. Bowman, Zac Hatfield-Dodds, Ben Mann, Dario Amodei, Nicholas Joseph, Sam McCandlish, Tom Brown, and Jared Kaplan.
\newblock Constitutional ai: Harmlessness from ai feedback.
\newblock 2022{\natexlab{b}}.

\bibitem[Cai et~al.(2024)Cai, Cao, Chen, Chen, Chen, Chen, Chen, Chen, Chen, Chu, et~al.]{cai2024internlm2}
Zheng Cai, Maosong Cao, Haojiong Chen, Kai Chen, Keyu Chen, Xin Chen, Xun Chen, Zehui Chen, Zhi Chen, Pei Chu, et~al.
\newblock Internlm2 technical report.
\newblock \emph{arXiv preprint arXiv:2403.17297}, 2024.

\bibitem[Chiang et~al.(2024)Chiang, Zheng, Sheng, Angelopoulos, Li, Li, Zhang, Zhu, Jordan, Gonzalez, and Stoica]{chiang2024chatbot}
Wei-Lin Chiang, Lianmin Zheng, Ying Sheng, Anastasios~Nikolas Angelopoulos, Tianle Li, Dacheng Li, Hao Zhang, Banghua Zhu, Michael Jordan, Joseph~E. Gonzalez, and Ion Stoica.
\newblock Chatbot arena: An open platform for evaluating llms by human preference, 2024.

\bibitem[Christiano et~al.(2023)Christiano, Leike, Brown, Martic, Legg, and Amodei]{christiano2023deep}
Paul Christiano, Jan Leike, Tom~B. Brown, Miljan Martic, Shane Legg, and Dario Amodei.
\newblock Deep reinforcement learning from human preferences.
\newblock 2023.

\bibitem[Dubois et~al.(2023)Dubois, Li, Taori, Zhang, Gulrajani, Ba, Guestrin, Liang, and Hashimoto]{dubois2023alpacafarm}
Yann Dubois, Xuechen Li, Rohan Taori, Tianyi Zhang, Ishaan Gulrajani, Jimmy Ba, Carlos Guestrin, Percy Liang, and Tatsunori~B Hashimoto.
\newblock Alpacafarm: A simulation framework for methods that learn from human feedback.
\newblock \emph{arXiv preprint arXiv:2305.14387}, 2023.

\bibitem[Ethayarajh et~al.(2022)Ethayarajh, Choi, and Swayamdipta]{pmlr-v162-ethayarajh22a}
Kawin Ethayarajh, Yejin Choi, and Swabha Swayamdipta.
\newblock Understanding dataset difficulty with $\mathcal{V}$-usable information.
\newblock In Kamalika Chaudhuri, Stefanie Jegelka, Le~Song, Csaba Szepesvari, Gang Niu, and Sivan Sabato (eds.), \emph{Proceedings of the 39th International Conference on Machine Learning}, volume 162 of \emph{Proceedings of Machine Learning Research}, pp.\  5988--6008. PMLR, 17--23 Jul 2022.
\newblock URL \url{https://proceedings.mlr.press/v162/ethayarajh22a.html}.

\bibitem[Frick et~al.(2024)Frick, Jin, Li, Ganesan, Zhang, Jiao, and Zhu]{athene70b2024}
Evan Frick, Peter Jin, Tianle Li, Karthik Ganesan, Jian Zhang, Jiantao Jiao, and Banghua Zhu.
\newblock Athene-70b: Redefining the boundaries of post-training for open models, July 2024.
\newblock URL \url{https://huggingface.co/Nexusflow/Athene-70B}.

\bibitem[Gao et~al.(2021)Gao, Tow, Biderman, Black, DiPofi, Foster, Golding, Hsu, McDonell, Muennighoff, Phang, Reynolds, Tang, Thite, Wang, Wang, and Zou]{eval-harness}
Leo Gao, Jonathan Tow, Stella Biderman, Sid Black, Anthony DiPofi, Charles Foster, Laurence Golding, Jeffrey Hsu, Kyle McDonell, Niklas Muennighoff, Jason Phang, Laria Reynolds, Eric Tang, Anish Thite, Ben Wang, Kevin Wang, and Andy Zou.
\newblock A framework for few-shot language model evaluation, September 2021.
\newblock URL \url{https://doi.org/10.5281/zenodo.5371628}.

\bibitem[Gao et~al.(2022)Gao, Schulman, and Hilton]{gao2022scaling}
Leo Gao, John Schulman, and Jacob Hilton.
\newblock Scaling laws for reward model overoptimization.
\newblock \emph{arXiv preprint arXiv:2210.10760}, 2022.

\bibitem[Hendrycks et~al.(2021)Hendrycks, Burns, Kadavath, Arora, Basart, Tang, Song, and Steinhardt]{hendrycksmath2021}
Dan Hendrycks, Collin Burns, Saurav Kadavath, Akul Arora, Steven Basart, Eric Tang, Dawn Song, and Jacob Steinhardt.
\newblock Measuring mathematical problem solving with the math dataset.
\newblock \emph{NeurIPS}, 2021.

\bibitem[Lambert et~al.(2024)Lambert, Pyatkin, Morrison, Miranda, Lin, Chandu, Dziri, Kumar, Zick, Choi, Smith, and Hajishirzi]{RewardBench}
Nathan Lambert, Valentina Pyatkin, Jacob Morrison, LJ~Miranda, Bill~Yuchen Lin, Khyathi Chandu, Nouha Dziri, Sachin Kumar, Tom Zick, Yejin Choi, Noah~A. Smith, and Hannaneh Hajishirzi.
\newblock Rewardbench: Evaluating reward models for language modeling.
\newblock \url{https://huggingface.co/spaces/allenai/reward-bench}, 2024.

\bibitem[Lee et~al.(2023)Lee, Phatale, Mansoor, Lu, Mesnard, Bishop, Carbune, and Rastogi]{lee2023rlaif}
Harrison Lee, Samrat Phatale, Hassan Mansoor, Kellie Lu, Thomas Mesnard, Colton Bishop, Victor Carbune, and Abhinav Rastogi.
\newblock Rlaif: Scaling reinforcement learning from human feedback with ai feedback.
\newblock 2023.

\bibitem[Li et~al.(2024{\natexlab{a}})Li, Angelopoulos, and Chiang]{stylearena2024}
Tianle Li, Anastasios Angelopoulos, and Wei-Lin Chiang.
\newblock Does style matter? disentangling style and substance in chatbot arena, August 2024{\natexlab{a}}.
\newblock URL \url{https://blog.lmarena.ai/blog/2024/style-control/}.

\bibitem[Li et~al.(2024{\natexlab{b}})Li, Chiang, Frick, Dunlap, Wu, Zhu, Gonzalez, and Stoica]{li2024crowdsourceddatahighqualitybenchmarks}
Tianle Li, Wei-Lin Chiang, Evan Frick, Lisa Dunlap, Tianhao Wu, Banghua Zhu, Joseph~E. Gonzalez, and Ion Stoica.
\newblock From crowdsourced data to high-quality benchmarks: Arena-hard and benchbuilder pipeline, 2024{\natexlab{b}}.
\newblock URL \url{https://arxiv.org/abs/2406.11939}.

\bibitem[Liu \& Zeng(2024)Liu and Zeng]{skyworkreward2024}
Chris~Yuhao Liu and Liang Zeng.
\newblock Skywork reward model series.
\newblock \url{https://huggingface.co/Skywork}, September 2024.
\newblock URL \url{https://huggingface.co/Skywork}.

\bibitem[OpenAI(2022)]{openai2022chatgpt}
OpenAI.
\newblock Introducing chatgpt.
\newblock \url{https://openai.com/blog/chatgpt}, 2022.
\newblock (Accessed on 01/12/2024).

\bibitem[OpenAI(2023{\natexlab{a}})]{openai2023gpt4}
OpenAI.
\newblock Gpt-4 technical report, 2023{\natexlab{a}}.

\bibitem[OpenAI(2023{\natexlab{b}})]{openai2023gpt4turbo}
OpenAI.
\newblock New models and developer products announced at devday.
\newblock \url{https://openai.com/blog/new-models-and-developer-products-announced-at-devday}, 2023{\natexlab{b}}.
\newblock (Accessed on 01/12/2024).

\bibitem[OpenAI(2024)]{openai2024gpt4omini}
OpenAI.
\newblock Gpt-4o mini: advancing cost-efficient intelligence.
\newblock \url{https://openai.com/index/gpt-4o-mini-advancing-cost-efficient-intelligence/}, 2024.
\newblock (Accessed on 06/05/2024).

\bibitem[Ouyang et~al.(2022)Ouyang, Wu, Jiang, Almeida, Wainwright, Mishkin, Zhang, Agarwal, Slama, Ray, et~al.]{ouyang2022training}
Long Ouyang, Jeffrey Wu, Xu~Jiang, Diogo Almeida, Carroll Wainwright, Pamela Mishkin, Chong Zhang, Sandhini Agarwal, Katarina Slama, Alex Ray, et~al.
\newblock Training language models to follow instructions with human feedback.
\newblock \emph{Advances in Neural Information Processing Systems}, 35:\penalty0 27730--27744, 2022.

\bibitem[Park et~al.(2024)Park, Jwa, Ren, Kim, and Choi]{park2024offsetbias}
Junsoo Park, Seungyeon Jwa, Meiying Ren, Daeyoung Kim, and Sanghyuk Choi.
\newblock Offsetbias: Leveraging debiased data for tuning evaluators, 2024.

\bibitem[Rafailov et~al.(2023)Rafailov, Sharma, Mitchell, Ermon, Manning, and Finn]{rafailov2023direct}
Rafael Rafailov, Archit Sharma, Eric Mitchell, Stefano Ermon, Christopher~D Manning, and Chelsea Finn.
\newblock Direct preference optimization: Your language model is secretly a reward model.
\newblock \emph{arXiv preprint arXiv:2305.18290}, 2023.

\bibitem[Rein et~al.(2023)Rein, Hou, Stickland, Petty, Pang, Dirani, Michael, and Bowman]{rein2023gpqa}
David Rein, Betty~Li Hou, Asa~Cooper Stickland, Jackson Petty, Richard~Yuanzhe Pang, Julien Dirani, Julian Michael, and Samuel~R. Bowman.
\newblock Gpqa: A graduate-level google-proof q\&a benchmark, 2023.

\bibitem[Schulman et~al.(2017)Schulman, Wolski, Dhariwal, Radford, and Klimov]{schulman2017proximal}
John Schulman, Filip Wolski, Prafulla Dhariwal, Alec Radford, and Oleg Klimov.
\newblock Proximal policy optimization algorithms.
\newblock \emph{arXiv preprint arXiv:1707.06347}, 2017.

\bibitem[Team et~al.(2024)Team, Riviere, Pathak, Sessa, Hardin, Bhupatiraju, Hussenot, Mesnard, Shahriari, Ram{\'e}, et~al.]{team2024gemma}
Gemma Team, Morgane Riviere, Shreya Pathak, Pier~Giuseppe Sessa, Cassidy Hardin, Surya Bhupatiraju, L{\'e}onard Hussenot, Thomas Mesnard, Bobak Shahriari, Alexandre Ram{\'e}, et~al.
\newblock Gemma 2: Improving open language models at a practical size.
\newblock \emph{arXiv preprint arXiv:2408.00118}, 2024.

\bibitem[Touvron et~al.(2023)Touvron, Lavril, Izacard, Martinet, Lachaux, Lacroix, Rozi{\`e}re, Goyal, Hambro, Azhar, et~al.]{touvron2023llama}
Hugo Touvron, Thibaut Lavril, Gautier Izacard, Xavier Martinet, Marie-Anne Lachaux, Timoth{\'e}e Lacroix, Baptiste Rozi{\`e}re, Naman Goyal, Eric Hambro, Faisal Azhar, et~al.
\newblock Llama: Open and efficient foundation language models.
\newblock \emph{arXiv preprint arXiv:2302.13971}, 2023.

\bibitem[von Werra et~al.(2020)von Werra, Belkada, Tunstall, Beeching, Thrush, Lambert, Huang, Rasul, and Gallouédec]{vonwerra2022trl}
Leandro von Werra, Younes Belkada, Lewis Tunstall, Edward Beeching, Tristan Thrush, Nathan Lambert, Shengyi Huang, Kashif Rasul, and Quentin Gallouédec.
\newblock Trl: Transformer reinforcement learning.
\newblock \url{https://github.com/huggingface/trl}, 2020.

\bibitem[Wang et~al.(2024{\natexlab{a}})Wang, Xiong, Xie, Zhao, and Zhang]{ArmoRM}
Haoxiang Wang, Wei Xiong, Tengyang Xie, Han Zhao, and Tong Zhang.
\newblock Interpretable preferences via multi-objective reward modeling and mixture-of-experts.
\newblock In \emph{EMNLP}, 2024{\natexlab{a}}.

\bibitem[Wang et~al.(2024{\natexlab{b}})Wang, Ma, Zhang, Ni, Chandra, Guo, Ren, Arulraj, He, Jiang, et~al.]{wang2024mmlu}
Yubo Wang, Xueguang Ma, Ge~Zhang, Yuansheng Ni, Abhranil Chandra, Shiguang Guo, Weiming Ren, Aaran Arulraj, Xuan He, Ziyan Jiang, et~al.
\newblock Mmlu-pro: A more robust and challenging multi-task language understanding benchmark.
\newblock \emph{arXiv preprint arXiv:2406.01574}, 2024{\natexlab{b}}.

\bibitem[Wang et~al.(2024{\natexlab{c}})Wang, Dong, Delalleau, Zeng, Shen, Egert, Zhang, Sreedhar, and Kuchaiev]{wang2024helpsteer2}
Zhilin Wang, Yi~Dong, Olivier Delalleau, Jiaqi Zeng, Gerald Shen, Daniel Egert, Jimmy~J. Zhang, Makesh~Narsimhan Sreedhar, and Oleksii Kuchaiev.
\newblock Helpsteer2: Open-source dataset for training top-performing reward models, 2024{\natexlab{c}}.

\bibitem[Yuan et~al.(2024)Yuan, Cui, Wang, Ding, Wang, Deng, Shan, Chen, Xie, Lin, Liu, Zhou, Peng, Liu, and Sun]{yuan2024advancing}
Lifan Yuan, Ganqu Cui, Hanbin Wang, Ning Ding, Xingyao Wang, Jia Deng, Boji Shan, Huimin Chen, Ruobing Xie, Yankai Lin, Zhenghao Liu, Bowen Zhou, Hao Peng, Zhiyuan Liu, and Maosong Sun.
\newblock Advancing llm reasoning generalists with preference trees, 2024.

\bibitem[Zeng et~al.(2024)Zeng, Yu, Gao, Meng, Goyal, and Chen]{zeng2024evaluatinglargelanguagemodels}
Zhiyuan Zeng, Jiatong Yu, Tianyu Gao, Yu~Meng, Tanya Goyal, and Danqi Chen.
\newblock Evaluating large language models at evaluating instruction following, 2024.
\newblock URL \url{https://arxiv.org/abs/2310.07641}.

\bibitem[Zheng et~al.(2023)Zheng, Chiang, Sheng, Zhuang, Wu, Zhuang, Lin, Li, Li, Xing, Zhang, Gonzalez, and Stoica]{zheng2023judging}
Lianmin Zheng, Wei-Lin Chiang, Ying Sheng, Siyuan Zhuang, Zhanghao Wu, Yonghao Zhuang, Zi~Lin, Zhuohan Li, Dacheng Li, Eric.~P Xing, Hao Zhang, Joseph~E. Gonzalez, and Ion Stoica.
\newblock Judging llm-as-a-judge with mt-bench and chatbot arena, 2023.

\bibitem[Zhou et~al.(2023)Zhou, Lu, Mishra, Brahma, Basu, Luan, Zhou, and Hou]{zhou2023instruction}
Jeffrey Zhou, Tianjian Lu, Swaroop Mishra, Siddhartha Brahma, Sujoy Basu, Yi~Luan, Denny Zhou, and Le~Hou.
\newblock Instruction-following evaluation for large language models.
\newblock \emph{arXiv preprint arXiv:2311.07911}, 2023.

\bibitem[Zhu et~al.(2023)Zhu, Frick, Wu, Zhu, and Jiao]{starling2023}
Banghua Zhu, Evan Frick, Tianhao Wu, Hanlin Zhu, and Jiantao Jiao.
\newblock Starling-7b: Improving llm helpfulness \& harmlessness with rlaif, November 2023.

\bibitem[Zhu et~al.(2024)Zhu, Frick, Wu, Zhu, Ganesan, Chiang, Zhang, and Jiao]{zhu2024starling}
Banghua Zhu, Evan Frick, Tianhao Wu, Hanlin Zhu, Karthik Ganesan, Wei-Lin Chiang, Jian Zhang, and Jiantao Jiao.
\newblock Starling-7b: Improving helpfulness and harmlessness with rlaif.
\newblock In \emph{First Conference on Language Modeling}, 2024.

\end{thebibliography}
\bibliographystyle{iclr2025_conference}
\newpage
\appendix
\section{Appendix}
\subsection{Detailed Scores for the Human Preference Evaluation Dataset}
\label{appendix:A}
You may include other additional sections here.

\begin{table}[H]
\centering
\resizebox{1\columnwidth}{!}{
\begin{tabular}{lrrrrrrr}
\toprule
Reward Model & Accuracy & R.W. Pearson & Separability & Conf. Agree. & Kendalltau & Spearmanr & Brier Score \\
\midrule
Ensemble-Judges (ArenaHard)\textsuperscript{\textdagger} & 68.59 & 82.49 & 84.21 & 96.21 & 87.37 & 96.54 & 0.05 \\
Ensemble-Judges (AlpacaEval)\textsuperscript{\textdagger} & 68.52 & 81.25 & 79.47 & 93.94 & 85.26 & 95.04 & 0.07 \\
GPT-4o-2024-08-06 (ArenaHard)\textsuperscript{\textdagger} & 67.71 & 81.07 & 80.53 & 94.70 & 86.32 & 96.24 & 0.06 \\
Claude-3-5-Sonnet-20240620 (ArenaHard)\textsuperscript{\textdagger} & 67.33 & 80.65 & 79.47 & 94.70 & 88.42 & 96.69 & 0.06 \\
GPT-4o-2024-08-06 (AlpacaEval)\textsuperscript{\textdagger} & 67.13 & 77.92 & 76.32 & 90.91 & 84.21 & 93.23 & 0.07 \\
Athene-RM-70B & 66.56 & 80.69 & 84.74 & 93.94 & 82.11 & 93.23 & 0.07 \\
GPT-4o-Mini-2024-07-18 (ArenaHard)\textsuperscript{\textdagger} & 66.46 & 78.42 & 75.26 & 92.42 & 83.16 & 93.08 & 0.07 \\
\bottomrule
\end{tabular}

}
\caption{Reward model and LLM judge performance on Overall subset of the human preference dataset. LLM-as-a-judge are labeled with system prompt source, and marked with \textdagger.}
\vspace{-0.5em}
\label{tab:overall_cap_table}
\end{table}

\begin{table}[H]
\centering
\resizebox{1\columnwidth}{!}{
\begin{tabular}{lrrrrrrr}
\toprule
Reward Model & Accuracy & R.W. Pearson & Separability & Conf. Agree. & Kendalltau & Spearmanr & Brier Score \\
\midrule
Ensemble-Judges (ArenaHard)\textsuperscript{\textdagger} & 69.46 & 67.05 & 74.21 & 96.88 & 83.16 & 94.44 & 0.06 \\
Claude-3-5-Sonnet-20240620 (ArenaHard)\textsuperscript{\textdagger} & 69.25 & 67.96 & 72.11 & 97.92 & 86.32 & 95.49 & 0.06 \\
GPT-4o-2024-08-06 (ArenaHard)\textsuperscript{\textdagger} & 68.50 & 68.17 & 71.05 & 97.92 & 85.26 & 95.94 & 0.06 \\
Ensemble-Judges (AlpacaEval)\textsuperscript{\textdagger} & 68.32 & 66.01 & 75.26 & 96.88 & 83.16 & 94.59 & 0.07 \\
GPT-4o-Mini-2024-07-18 (ArenaHard)\textsuperscript{\textdagger} & 66.63 & 63.55 & 71.05 & 95.83 & 82.11 & 94.29 & 0.08 \\
Gemini-1.5-Pro-002 (AlpacaEval)\textsuperscript{\textdagger} & 66.53 & 66.85 & 72.63 & 96.88 & 84.21 & 95.49 & 0.06 \\
Athene-RM-70B & 66.43 & 67.01 & 76.84 & 96.88 & 78.95 & 92.93 & 0.08 \\
\bottomrule
\end{tabular}

}
\caption{Reward model and LLM judge performance on Hard prompt subset of the human preference dataset. LLM-as-a-judge are labeled with system prompt source, and marked with \textdagger.}
\vspace{-0.5em}
\label{tab:hard_prompt_cap_table}
\end{table}

\begin{table}[H]
\centering
\resizebox{1\columnwidth}{!}{
\begin{tabular}{lrrrrrrr}
\toprule
Reward Model & Accuracy & R.W. Pearson & Separability & Conf. Agree. & Kendalltau & Spearmanr & Brier Score \\
\midrule
Ensemble-Judges (AlpacaEval)\textsuperscript{\textdagger} & 70.15 & 52.24 & 52.10 & 83.33 & 75.79 & 91.58 & 0.09 \\
GPT-4o-2024-08-06 (AlpacaEval)\textsuperscript{\textdagger} & 69.97 & 52.01 & 47.37 & 83.33 & 72.63 & 90.08 & 0.09 \\
Ensemble-Judges (ArenaHard)\textsuperscript{\textdagger} & 69.59 & 57.24 & 63.16 & 83.33 & 83.16 & 94.74 & 0.08 \\
GPT-4o-2024-08-06 (ArenaHard)\textsuperscript{\textdagger} & 68.54 & 56.01 & 52.10 & 81.25 & 77.89 & 93.53 & 0.08 \\
GPT-4o-Mini-2024-07-18 (ArenaHard)\textsuperscript{\textdagger} & 67.50 & 50.08 & 46.32 & 78.12 & 72.63 & 88.72 & 0.09 \\
Llama-3.1-70B-Instruct (AlpacaEval)\textsuperscript{\textdagger} & 67.40 & 46.25 & 46.32 & 68.75 & 60.00 & 80.60 & 0.14 \\
Gemini-1.5-Pro-002 (ArenaHard)\textsuperscript{\textdagger} & 67.08 & 55.16 & 57.37 & 90.62 & 82.11 & 94.89 & 0.06 \\
\bottomrule
\end{tabular}

}
\caption{Reward model and LLM judge performance on Easy prompt subset of the human preference dataset. LLM-as-a-judge are labeled with system prompt source, and marked with \textdagger.}
\vspace{-0.5em}
\label{tab:easy_prompt_cap_table}
\end{table}

\begin{table}[H]
\centering
\resizebox{1\columnwidth}{!}{
\begin{tabular}{lrrrrrrr}
\toprule
Reward Model & Accuracy & R.W. Pearson & Separability & Conf. Agree. & Kendalltau & Spearmanr & Brier Score \\
\midrule
Ensemble-Judges (ArenaHard)\textsuperscript{\textdagger} & 69.77 & 66.89 & 70.00 & 97.09 & 83.16 & 93.68 & 0.07 \\
Claude-3-5-Sonnet-20240620 (ArenaHard)\textsuperscript{\textdagger} & 68.38 & 70.13 & 64.74 & 92.23 & 80.00 & 91.88 & 0.07 \\
Ensemble-Judges (AlpacaEval)\textsuperscript{\textdagger} & 67.86 & 69.18 & 70.00 & 96.12 & 86.32 & 95.04 & 0.05 \\
GPT-4o-2024-08-06 (ArenaHard)\textsuperscript{\textdagger} & 67.51 & 60.99 & 66.84 & 96.12 & 78.95 & 92.93 & 0.08 \\
Gemini-1.5-Pro-002 (AlpacaEval)\textsuperscript{\textdagger} & 66.78 & 68.61 & 73.16 & 97.09 & 88.42 & 96.54 & 0.04 \\
Gemini-1.5-Pro-002 (ArenaHard)\textsuperscript{\textdagger} & 66.70 & 69.92 & 68.42 & 97.09 & 82.11 & 93.83 & 0.06 \\
Athene-RM-70B & 66.50 & 63.79 & 75.26 & 95.15 & 77.89 & 90.98 & 0.09 \\
\bottomrule
\end{tabular}

}
\caption{Reward model and LLM judge performance on If prompt subset of the human preference dataset. LLM-as-a-judge are labeled with system prompt source, and marked with \textdagger.}
\vspace{-0.5em}
\label{tab:if_prompt_cap_table}
\end{table}

\begin{table}[H]
\centering
\resizebox{1\columnwidth}{!}{
\begin{tabular}{lrrrrrrr}
\toprule
Reward Model & Accuracy & R.W. Pearson & Separability & Conf. Agree. & Kendalltau & Spearmanr & Brier Score \\
\midrule
Claude-3-5-Sonnet-20240620 (ArenaHard)\textsuperscript{\textdagger} & 68.06 & 57.64 & 62.63 & 97.22 & 88.42 & 97.74 & 0.04 \\
Ensemble-Judges (ArenaHard)\textsuperscript{\textdagger} & 67.98 & 58.22 & 71.58 & 91.67 & 84.21 & 96.09 & 0.05 \\
GPT-4o-2024-08-06 (ArenaHard)\textsuperscript{\textdagger} & 67.66 & 58.16 & 65.79 & 97.22 & 88.42 & 97.29 & 0.04 \\
Ensemble-Judges (AlpacaEval)\textsuperscript{\textdagger} & 67.47 & 55.98 & 72.11 & 94.44 & 82.11 & 94.14 & 0.06 \\
Athene-RM-70B & 66.87 & 57.57 & 70.53 & 94.44 & 81.05 & 93.23 & 0.07 \\
GPT-4o-Mini-2024-07-18 (ArenaHard)\textsuperscript{\textdagger} & 66.08 & 53.90 & 67.90 & 100.00 & 85.26 & 96.24 & 0.05 \\
Claude-3-5-Sonnet-20240620 (AlpacaEval)\textsuperscript{\textdagger} & 65.92 & 45.70 & 60.00 & 97.22 & 81.05 & 94.44 & 0.08 \\
\bottomrule
\end{tabular}

}
\caption{Reward model and LLM judge performance on Is code subset of the human preference dataset. LLM-as-a-judge are labeled with system prompt source, and marked with \textdagger.}
\vspace{-0.5em}
\label{tab:is_code_cap_table}
\end{table}

\begin{table}[H]
\centering
\resizebox{1\columnwidth}{!}{
\begin{tabular}{lrrrrrrr}
\toprule
Reward Model & Accuracy & R.W. Pearson & Separability & Conf. Agree. & Kendalltau & Spearmanr & Brier Score \\
\midrule
Ensemble-Judges (ArenaHard)\textsuperscript{\textdagger} & 73.58 & 54.87 & 65.79 & 88.73 & 80.00 & 94.44 & 0.07 \\
GPT-4o-2024-08-06 (ArenaHard)\textsuperscript{\textdagger} & 72.57 & 56.46 & 63.16 & 88.73 & 82.11 & 94.89 & 0.06 \\
Claude-3-5-Sonnet-20240620 (ArenaHard)\textsuperscript{\textdagger} & 71.79 & 49.92 & 60.53 & 88.73 & 78.95 & 93.38 & 0.08 \\
GPT-4o-Mini-2024-07-18 (ArenaHard)\textsuperscript{\textdagger} & 70.20 & 50.30 & 55.26 & 87.32 & 71.58 & 87.97 & 0.11 \\
Gemini-1.5-Pro-002 (ArenaHard)\textsuperscript{\textdagger} & 69.61 & 60.91 & 58.42 & 84.51 & 77.89 & 92.63 & 0.08 \\
Ensemble-Judges (AlpacaEval)\textsuperscript{\textdagger} & 69.09 & 52.15 & 62.10 & 91.55 & 74.74 & 91.13 & 0.09 \\
Llama-3.1-70B-Instruct (ArenaHard)\textsuperscript{\textdagger} & 68.93 & 46.05 & 54.74 & 84.51 & 72.63 & 87.82 & 0.10 \\
\bottomrule
\end{tabular}

}
\caption{Reward model and LLM judge performance on Math prompt subset of the human preference dataset. LLM-as-a-judge are labeled with system prompt source, and marked with \textdagger.}
\vspace{-0.5em}
\label{tab:math_prompt_cap_table}
\end{table}

\begin{table}[H]
\centering
\resizebox{1\columnwidth}{!}{
\begin{tabular}{lrrrrrrr}
\toprule
Reward Model & Accuracy & R.W. Pearson & Separability & Conf. Agree. & Kendalltau & Spearmanr & Brier Score \\
\midrule
Nemotron-4-340B-Reward & 62.65 & 56.88 & 58.95 & 62.28 & 51.58 & 68.42 & 0.19 \\
Gemini-1.5-Pro-002 (ArenaHard)\textsuperscript{\textdagger} & 59.90 & 45.67 & 66.32 & 44.74 & 37.89 & 53.38 & 0.27 \\
Gemini-1.5-Pro-001 (ArenaHard)\textsuperscript{\textdagger} & 58.01 & 36.29 & 52.63 & 42.11 & 41.05 & 53.23 & 0.27 \\
ArmoRM-Llama3-8B-v0.1 & 56.83 & 33.59 & 43.16 & 42.98 & 36.84 & 47.82 & 0.27 \\
Gemini-1.5-Pro-002 (AlpacaEval)\textsuperscript{\textdagger} & 56.83 & 30.75 & 67.90 & 38.60 & 30.53 & 45.41 & 0.31 \\
Athene-RM-70B & 55.81 & 31.06 & 67.37 & 35.96 & 28.42 & 44.06 & 0.32 \\
Ensemble-Judges (ArenaHard)\textsuperscript{\textdagger} & 55.27 & 36.57 & 66.32 & 42.11 & 37.89 & 53.68 & 0.27 \\
\bottomrule
\end{tabular}

}
\caption{Reward model and LLM judge performance on Shorter won subset of the human preference dataset. LLM-as-a-judge are labeled with system prompt source, and marked with \textdagger.}
\vspace{-0.5em}
\label{tab:shorter_won_cap_table}
\end{table}

\begin{table}[H]
\centering
\resizebox{1\columnwidth}{!}{
\begin{tabular}{lrrrrrrr}
\toprule
Reward Model & Accuracy & R.W. Pearson & Separability & Conf. Agree. & Kendalltau & Spearmanr & Brier Score \\
\midrule
Ensemble-Judges (ArenaHard)\textsuperscript{\textdagger} & 68.15 & 71.49 & 73.16 & 91.59 & 86.32 & 95.64 & 0.06 \\
Ensemble-Judges (AlpacaEval)\textsuperscript{\textdagger} & 67.28 & 73.31 & 74.21 & 92.52 & 84.21 & 94.44 & 0.06 \\
GPT-4o-2024-08-06 (ArenaHard)\textsuperscript{\textdagger} & 67.23 & 71.93 & 71.05 & 92.52 & 84.21 & 95.19 & 0.07 \\
Claude-3-5-Sonnet-20240620 (ArenaHard)\textsuperscript{\textdagger} & 67.08 & 72.22 & 70.00 & 88.79 & 84.21 & 93.83 & 0.06 \\
GPT-4o-Mini-2024-07-18 (ArenaHard)\textsuperscript{\textdagger} & 66.29 & 71.23 & 69.47 & 89.72 & 80.00 & 92.48 & 0.08 \\
Athene-RM-70B & 65.84 & 72.39 & 81.05 & 90.65 & 78.95 & 91.88 & 0.09 \\
Gemini-1.5-Pro-002 (AlpacaEval)\textsuperscript{\textdagger} & 65.54 & 71.75 & 74.21 & 92.52 & 85.26 & 94.74 & 0.06 \\
\bottomrule
\end{tabular}

}
\caption{Reward model and LLM judge performance on Similar response subset of the human preference dataset. LLM-as-a-judge are labeled with system prompt source, and marked with \textdagger.}
\vspace{-0.5em}
\label{tab:similar_response_cap_table}
\end{table}

\begin{table}[H]
\centering
\resizebox{1\columnwidth}{!}{
\begin{tabular}{lrrrrrrr}
\toprule
Reward Model & Accuracy & R.W. Pearson & Separability & Conf. Agree. & Kendalltau & Spearmanr & Brier Score \\
\midrule
Ensemble-Judges (ArenaHard)\textsuperscript{\textdagger} & 68.17 & 70.80 & 71.58 & 86.24 & 81.05 & 94.14 & 0.08 \\
GPT-4o-2024-08-06 (ArenaHard)\textsuperscript{\textdagger} & 67.78 & 71.61 & 68.95 & 86.24 & 83.16 & 94.89 & 0.07 \\
Ensemble-Judges (AlpacaEval)\textsuperscript{\textdagger} & 67.60 & 70.66 & 71.58 & 84.40 & 76.84 & 92.93 & 0.10 \\
GPT-4o-2024-08-06 (AlpacaEval)\textsuperscript{\textdagger} & 66.70 & 63.51 & 66.32 & 80.73 & 76.84 & 91.73 & 0.09 \\
Claude-3-5-Sonnet-20240620 (ArenaHard)\textsuperscript{\textdagger} & 66.42 & 68.25 & 70.53 & 86.24 & 78.95 & 93.68 & 0.08 \\
GPT-4o-Mini-2024-07-18 (ArenaHard)\textsuperscript{\textdagger} & 66.39 & 66.39 & 67.37 & 81.65 & 78.95 & 92.03 & 0.09 \\
Athene-RM-70B & 65.53 & 68.75 & 79.47 & 83.49 & 73.68 & 90.98 & 0.12 \\
\bottomrule
\end{tabular}

}
\caption{Reward model and LLM judge performance on English prompt subset of the human preference dataset. LLM-as-a-judge are labeled with system prompt source, and marked with \textdagger.}
\vspace{-0.5em}
\label{tab:english_prompt_cap_table}
\end{table}

\begin{table}[H]
\centering
\resizebox{1\columnwidth}{!}{
\begin{tabular}{lrrrrrrr}
\toprule
Reward Model & Accuracy & R.W. Pearson & Separability & Conf. Agree. & Kendalltau & Spearmanr & Brier Score \\
\midrule
Ensemble-Judges (AlpacaEval)\textsuperscript{\textdagger} & 69.68 & 73.76 & 74.21 & 94.31 & 90.53 & 97.74 & 0.03 \\
Ensemble-Judges (ArenaHard)\textsuperscript{\textdagger} & 69.09 & 75.81 & 76.84 & 93.50 & 86.32 & 95.79 & 0.06 \\
Claude-3-5-Sonnet-20240620 (ArenaHard)\textsuperscript{\textdagger} & 68.48 & 75.18 & 75.26 & 91.87 & 86.32 & 96.39 & 0.05 \\
Athene-RM-70B & 67.86 & 73.24 & 76.84 & 91.87 & 82.11 & 94.89 & 0.07 \\
GPT-4o-2024-08-06 (AlpacaEval)\textsuperscript{\textdagger} & 67.66 & 72.18 & 72.63 & 98.37 & 93.68 & 98.65 & 0.03 \\
GPT-4o-2024-08-06 (ArenaHard)\textsuperscript{\textdagger} & 67.63 & 71.24 & 73.16 & 91.87 & 82.11 & 94.74 & 0.07 \\
Gemini-1.5-Pro-002 (AlpacaEval)\textsuperscript{\textdagger} & 67.01 & 73.72 & 80.00 & 94.31 & 88.42 & 97.14 & 0.05 \\
\bottomrule
\end{tabular}

}
\caption{Reward model and LLM judge performance on Non english prompt subset of the human preference dataset. LLM-as-a-judge are labeled with system prompt source, and marked with \textdagger.}
\vspace{-0.5em}
\label{tab:non_english_prompt_cap_table}
\end{table}

\begin{table}[H]
\centering
\resizebox{1\columnwidth}{!}{
\begin{tabular}{lrrrrrrr}
\toprule
Reward Model & Accuracy & R.W. Pearson & Separability & Conf. Agree. & Kendalltau & Spearmanr & Brier Score \\
\midrule
Ensemble-Judges (AlpacaEval)\textsuperscript{\textdagger} & 67.91 & 52.67 & 54.21 & 93.33 & 80.00 & 94.14 & 0.07 \\
Claude-3-5-Sonnet-20240620 (ArenaHard)\textsuperscript{\textdagger} & 67.03 & 50.91 & 48.42 & 90.00 & 78.95 & 93.38 & 0.08 \\
Athene-RM-70B & 66.39 & 45.24 & 61.05 & 90.00 & 83.16 & 93.83 & 0.07 \\
Gemini-1.5-Pro-002 (AlpacaEval)\textsuperscript{\textdagger} & 66.27 & 49.83 & 58.42 & 93.33 & 82.11 & 93.38 & 0.08 \\
Ensemble-Judges (ArenaHard)\textsuperscript{\textdagger} & 66.15 & 53.77 & 47.37 & 86.67 & 77.89 & 92.33 & 0.07 \\
GPT-4o-2024-08-06 (ArenaHard)\textsuperscript{\textdagger} & 65.37 & 49.18 & 52.10 & 90.00 & 76.84 & 92.18 & 0.08 \\
GPT-4o-Mini-2024-07-18 (ArenaHard)\textsuperscript{\textdagger} & 65.29 & 51.87 & 44.74 & 76.67 & 66.32 & 86.47 & 0.12 \\
\bottomrule
\end{tabular}

}
\caption{Reward model and LLM judge performance on Chinese prompt subset of the human preference dataset. LLM-as-a-judge are labeled with system prompt source, and marked with \textdagger.}
\vspace{-0.5em}
\label{tab:chinese_prompt_cap_table}
\end{table}

\begin{table}[H]
\centering
\resizebox{1\columnwidth}{!}{
\begin{tabular}{lrrrrrrr}
\toprule
Reward Model & Accuracy & R.W. Pearson & Separability & Conf. Agree. & Kendalltau & Spearmanr & Brier Score \\
\midrule
Ensemble-Judges (ArenaHard)\textsuperscript{\textdagger} & 70.37 & 50.61 & 53.16 & 92.86 & 77.89 & 92.63 & 0.09 \\
Ensemble-Judges (AlpacaEval)\textsuperscript{\textdagger} & 69.43 & 51.76 & 57.90 & 92.86 & 80.00 & 94.44 & 0.06 \\
Claude-3-5-Sonnet-20240620 (ArenaHard)\textsuperscript{\textdagger} & 68.63 & 44.71 & 50.53 & 85.71 & 70.53 & 87.97 & 0.09 \\
GPT-4o-2024-08-06 (AlpacaEval)\textsuperscript{\textdagger} & 68.58 & 42.94 & 38.95 & 91.07 & 77.89 & 93.83 & 0.08 \\
GPT-4o-2024-08-06 (ArenaHard)\textsuperscript{\textdagger} & 68.54 & 43.94 & 47.37 & 89.29 & 70.53 & 89.02 & 0.10 \\
Athene-RM-70B & 68.49 & 48.66 & 58.42 & 94.64 & 77.89 & 90.68 & 0.09 \\
Gemini-1.5-Pro-002 (ArenaHard)\textsuperscript{\textdagger} & 67.23 & 49.82 & 53.68 & 87.50 & 73.68 & 89.32 & 0.10 \\
\bottomrule
\end{tabular}

}
\caption{Reward model and LLM judge performance on Russian prompt subset of the human preference dataset. LLM-as-a-judge are labeled with system prompt source, and marked with \textdagger.}
\vspace{-0.5em}
\label{tab:russian_prompt_cap_table}
\end{table}

\begin{table}[H]
\centering
\resizebox{1\columnwidth}{!}{
\begin{tabular}{lrrrrrrr}
\toprule
Reward Model & Accuracy & R.W. Pearson & Separability & Conf. Agree. & Kendalltau & Spearmanr & Brier Score \\
\midrule
Ensemble-Judges (ArenaHard)\textsuperscript{\textdagger} & 75.16 & 38.73 & 38.42 & 84.62 & 73.68 & 88.42 & 0.10 \\
Claude-3-5-Sonnet-20240620 (ArenaHard)\textsuperscript{\textdagger} & 72.49 & 30.32 & 23.16 & 66.67 & 65.26 & 81.50 & 0.12 \\
GPT-4o-2024-08-06 (ArenaHard)\textsuperscript{\textdagger} & 71.03 & 31.32 & 24.74 & 84.62 & 72.63 & 85.86 & 0.10 \\
Gemini-1.5-Pro-002 (ArenaHard)\textsuperscript{\textdagger} & 70.64 & 29.57 & 27.89 & 76.92 & 72.63 & 87.22 & 0.11 \\
GPT-4o-2024-08-06 (AlpacaEval)\textsuperscript{\textdagger} & 69.71 & 21.47 & 21.05 & 74.36 & 72.63 & 88.27 & 0.10 \\
Ensemble-Judges (AlpacaEval)\textsuperscript{\textdagger} & 68.88 & 15.78 & 27.37 & 71.79 & 60.00 & 78.05 & 0.14 \\
Athene-RM-70B & 67.71 & 11.39 & 33.68 & 76.92 & 65.26 & 84.21 & 0.13 \\
\bottomrule
\end{tabular}

}
\caption{Reward model and LLM judge performance on German prompt subset of the human preference dataset. LLM-as-a-judge are labeled with system prompt source, and marked with \textdagger.}
\vspace{-0.5em}
\label{tab:german_prompt_cap_table}
\end{table}

\begin{table}[H]
\centering
\resizebox{1\columnwidth}{!}{
\begin{tabular}{lrrrrrrr}
\toprule
Reward Model & Accuracy & R.W. Pearson & Separability & Conf. Agree. & Kendalltau & Spearmanr & Brier Score \\
\midrule
Athene-RM-70B & 71.10 & 46.16 & 37.37 & 84.21 & 67.37 & 83.76 & 0.14 \\
Ensemble-Judges (AlpacaEval)\textsuperscript{\textdagger} & 69.63 & 32.44 & 34.21 & 52.63 & 63.16 & 82.71 & 0.13 \\
Skywork-Reward-Llama-3.1-8B & 68.81 & 40.32 & 22.11 & 68.42 & 58.95 & 78.20 & 0.14 \\
Ensemble-Judges (ArenaHard)\textsuperscript{\textdagger} & 68.45 & 33.85 & 25.79 & 65.79 & 61.05 & 78.50 & 0.14 \\
Gemini-1.5-Pro-002 (AlpacaEval)\textsuperscript{\textdagger} & 68.06 & 28.63 & 28.42 & 50.00 & 66.32 & 84.36 & 0.12 \\
Claude-3-5-Sonnet-20240620 (AlpacaEval)\textsuperscript{\textdagger} & 67.59 & 27.29 & 12.11 & 36.84 & 57.89 & 78.95 & 0.15 \\
Llama-3.1-70B-Instruct (AlpacaEval)\textsuperscript{\textdagger} & 66.97 & 24.59 & 19.47 & 52.63 & 61.05 & 78.20 & 0.15 \\
\bottomrule
\end{tabular}

}
\caption{Reward model and LLM judge performance on Korean prompt subset of the human preference dataset. LLM-as-a-judge are labeled with system prompt source, and marked with \textdagger.}
\vspace{-0.5em}
\label{tab:korean_prompt_cap_table}
\end{table}

\begin{table}[H]
\centering
\resizebox{1\columnwidth}{!}{
\begin{tabular}{lrrrrrrr}
\toprule
Reward Model & Accuracy & R.W. Pearson & Separability & Conf. Agree. & Kendalltau & Spearmanr & Brier Score \\
\midrule
Claude-3-5-Sonnet-20240620 (AlpacaEval)\textsuperscript{\textdagger} & 73.36 & 37.78 & 6.32 & 58.33 & 69.47 & 87.22 & 0.11 \\
Athene-RM-8B & 71.89 & 39.72 & 14.21 & 54.17 & 67.37 & 87.07 & 0.10 \\
Ensemble-Judges (AlpacaEval)\textsuperscript{\textdagger} & 71.36 & 36.61 & 11.05 & 70.83 & 71.58 & 86.62 & 0.11 \\
Llama-3.1-70B-Instruct (AlpacaEval)\textsuperscript{\textdagger} & 70.05 & 37.95 & 6.32 & 62.50 & 62.11 & 81.50 & 0.11 \\
Claude-3-5-Sonnet-20240620 (ArenaHard)\textsuperscript{\textdagger} & 68.52 & 33.33 & 14.74 & 75.00 & 72.63 & 89.62 & 0.10 \\
Athene-RM-70B & 68.20 & 33.11 & 18.42 & 50.00 & 72.63 & 87.82 & 0.13 \\
GPT-4o-Mini-2024-07-18 (ArenaHard)\textsuperscript{\textdagger} & 68.20 & 41.02 & 8.95 & 58.33 & 62.11 & 80.75 & 0.13 \\
\bottomrule
\end{tabular}

}
\caption{Reward model and LLM judge performance on Japanese prompt subset of the human preference dataset. LLM-as-a-judge are labeled with system prompt source, and marked with \textdagger.}
\vspace{-0.5em}
\label{tab:japanese_prompt_cap_table}
\end{table}

\begin{table}[H]
\centering
\resizebox{1\columnwidth}{!}{
\begin{tabular}{lrrrrrrr}
\toprule
Reward Model & Accuracy & R.W. Pearson & Separability & Conf. Agree. & Kendalltau & Spearmanr & Brier Score \\
\midrule
Ensemble-Judges (AlpacaEval)\textsuperscript{\textdagger} & 72.11 & 31.81 & 5.79 & 36.84 & 20.00 & 30.53 & 0.28 \\
GPT-4o-2024-08-06 (AlpacaEval)\textsuperscript{\textdagger} & 70.53 & 23.71 & 0.00 & 100.00 & 35.79 & 48.42 & 0.22 \\
GPT-4o-2024-08-06 (ArenaHard)\textsuperscript{\textdagger} & 70.29 & 24.79 & 4.21 & 89.47 & 43.16 & 59.55 & 0.21 \\
Athene-RM-70B & 69.47 & 24.25 & 17.37 & 89.47 & 35.79 & 49.62 & 0.23 \\
Claude-3-5-Sonnet-20240620 (AlpacaEval)\textsuperscript{\textdagger} & 68.42 & 28.53 & 1.58 & 100.00 & 20.00 & 33.83 & 0.26 \\
Llama-3.1-70B-Instruct (ArenaHard)\textsuperscript{\textdagger} & 67.93 & 29.52 & 6.32 & 78.95 & 25.26 & 32.63 & 0.28 \\
Skywork-Reward-Llama-3.1-8B & 67.89 & 20.95 & 7.37 & 89.47 & 35.79 & 52.33 & 0.21 \\
\bottomrule
\end{tabular}

}
\caption{Reward model and LLM judge performance on Spanish prompt subset of the human preference dataset. LLM-as-a-judge are labeled with system prompt source, and marked with \textdagger.}
\vspace{-0.5em}
\label{tab:spanish_prompt_cap_table}
\end{table}

\begin{table}[H]
\centering
\resizebox{1\columnwidth}{!}{
\begin{tabular}{lrrrrrrr}
\toprule
Reward Model & Accuracy & R.W. Pearson & Separability & Conf. Agree. & Kendalltau & Spearmanr & Brier Score \\
\midrule
Gemini-1.5-Pro-002 (ArenaHard)\textsuperscript{\textdagger} & 69.57 & 14.77 & 14.74 & 54.17 & 63.16 & 82.41 & 0.14 \\
GPT-4o-Mini-2024-07-18 (ArenaHard)\textsuperscript{\textdagger} & 68.45 & 25.12 & 4.21 & 75.00 & 54.74 & 73.08 & 0.17 \\
Ensemble-Judges (ArenaHard)\textsuperscript{\textdagger} & 68.24 & 21.05 & 17.37 & 66.67 & 62.11 & 80.90 & 0.13 \\
Ensemble-Judges (AlpacaEval)\textsuperscript{\textdagger} & 67.74 & 27.12 & 4.21 & 79.17 & 46.32 & 65.71 & 0.19 \\
Gemini-1.5-Pro-002 (AlpacaEval)\textsuperscript{\textdagger} & 67.38 & 26.42 & 8.95 & 79.17 & 47.37 & 65.26 & 0.18 \\
Athene-RM-8B & 67.38 & 26.84 & 18.95 & 45.83 & 45.26 & 64.81 & 0.17 \\
InternLM2-7B-Reward & 66.31 & 20.42 & 11.05 & 45.83 & 43.16 & 62.41 & 0.19 \\
\bottomrule
\end{tabular}

}
\caption{Reward model and LLM judge performance on French prompt subset of the human preference dataset. LLM-as-a-judge are labeled with system prompt source, and marked with \textdagger.}
\vspace{-0.5em}
\label{tab:french_prompt_cap_table}
\end{table}

\begin{table}[H]
\centering
\resizebox{1\columnwidth}{!}{
\begin{tabular}{lrrrrrrr}
\toprule
Reward Model & Accuracy & R.W. Pearson & Separability & Conf. Agree. & Kendalltau & Spearmanr & Brier Score \\
\midrule
GPT-4o-Mini-2024-07-18 (AlpacaEval)\textsuperscript{\textdagger} & 71.84 & 31.95 & 2.11 & 100.00 & 49.47 & 67.82 & 0.18 \\
Claude-3-5-Sonnet-20240620 (AlpacaEval)\textsuperscript{\textdagger} & 68.93 & 27.08 & 7.37 & 100.00 & 48.42 & 67.97 & 0.22 \\
InternLM2-7B-Reward & 68.93 & 25.47 & 1.05 & 100.00 & 49.47 & 68.12 & 0.18 \\
Claude-3-5-Sonnet-20240620 (ArenaHard)\textsuperscript{\textdagger} & 68.63 & 20.55 & 3.68 & 100.00 & 60.00 & 77.74 & 0.18 \\
Ensemble-Judges (AlpacaEval)\textsuperscript{\textdagger} & 67.96 & 17.35 & 7.37 & 100.00 & 57.89 & 79.25 & 0.22 \\
Ensemble-Judges (ArenaHard)\textsuperscript{\textdagger} & 67.02 & 20.72 & 10.53 & 100.00 & 62.11 & 76.39 & 0.17 \\
GPT-4o-2024-08-06 (AlpacaEval)\textsuperscript{\textdagger} & 66.99 & 16.25 & 3.68 & 100.00 & 50.53 & 69.47 & 0.18 \\
\bottomrule
\end{tabular}

}
\caption{Reward model and LLM judge performance on Portuguese prompt subset of the human preference dataset. LLM-as-a-judge are labeled with system prompt source, and marked with \textdagger.}
\vspace{-0.5em}
\label{tab:portuguese_prompt_cap_table}
\end{table}

\begin{table}[H]
\centering
\resizebox{1\columnwidth}{!}{
\begin{tabular}{lrrrrrrr}
\toprule
Reward Model & Accuracy & R.W. Pearson & Separability & Conf. Agree. & Kendalltau & Spearmanr & Brier Score \\
\midrule
Gemini-1.5-Pro-002 (AlpacaEval)\textsuperscript{\textdagger} & 81.40 & 51.04 & 3.16 & 100.00 & 50.53 & 74.14 & 0.17 \\
Ensemble-Judges (AlpacaEval)\textsuperscript{\textdagger} & 75.58 & 44.04 & 6.84 & 100.00 & 45.26 & 66.47 & 0.18 \\
Gemini-1.5-Pro-002 (ArenaHard)\textsuperscript{\textdagger} & 74.42 & 40.23 & 3.16 & 57.89 & 52.63 & 70.83 & 0.18 \\
Athene-RM-70B & 74.42 & 42.65 & 4.74 & 100.00 & 43.16 & 61.05 & 0.20 \\
Claude-3-5-Sonnet-20240620 (ArenaHard)\textsuperscript{\textdagger} & 73.26 & 42.33 & 1.58 & 100.00 & 47.37 & 58.80 & 0.20 \\
Athene-RM-8B & 73.26 & 43.29 & 8.42 & 78.95 & 43.16 & 60.45 & 0.19 \\
Ensemble-Judges (ArenaHard)\textsuperscript{\textdagger} & 71.25 & 44.59 & 1.58 & 89.47 & 36.84 & 51.88 & 0.20 \\
\bottomrule
\end{tabular}

}
\caption{Reward model and LLM judge performance on Italian prompt subset of the human preference dataset. LLM-as-a-judge are labeled with system prompt source, and marked with \textdagger.}
\vspace{-0.5em}
\label{tab:italian_prompt_cap_table}
\end{table}

\subsection{Details on Curation and Scores for Correctness Preference Evaluation Dataset} \label{appendix: B}

\subsubsection{Small Benchmark Modifications} \label{appendix:smalle-bench-mod}


To ensure more natural responses that better reflect real-world use cases, we modified each verifiable benchmark's canonical prompt to encourage Chain of Thought (CoT) thinking (citation). This approach both increases the diversity of sampled responses and enhances the task difficulty for the human preference proxy by incorporating additional signals beyond final answer correctness. The specific instructions for each benchmark are detailed below.


For the MATH benchmark, we implemented a new system prompt to facilitate zero-shot CoT behavior. Additionally, we converted the parsed answer to its symbolic representation and utilized a symbolic solver to evaluate true equality instead of relying on raw string matching. This refinement of the correctness signal ensures that trivial answer differences, such as $1\frac{3}{4}$ vs $\frac{7}{4}$ or $\frac{4i + \sqrt{5}}{2}$ vs $\frac{\sqrt{5}}{2} + 2i$, are marked as equivalent, with either answer accepted if correct.


In practice, we observed that the sampled MBPP-Plus generations from some models were almost all identical. Models also generally failed to follow instructions to ``think step-by-step" before providing their final answers, suppressing answer diversity. To address this issue, we prompted the models to ``write comments clearly explaining each part of the code," thereby lengthening trajectories and yielding greater exploration of the answer spaces. We also observed some ambiguity in MBPP-Plus intructions. To mitigate this, we added standard MBPP test cases into the function docstring as examples, and used the more extensive remaining MBPP-Plus test cases as the real tests.


Lastly, for IFEval, we prefixed the prompts with ``It is extremely important that you follow all instructions exactly." This addition emphasizes the necessity of precise instruction following in these tasks and ensures that the human preference proxy implicitly recognizes this as a significant evaluation criterion.

The prompt template for MMLU-Pro and GPQA were adaption from \citet{eval-harness}'s Language Model Evaluation Harness. The MATH template was generated with the assistance of Anthropic's prompt generator.

The prompt templates for each benchmark are detailed below. Note that \texttt{\{\{var\}\}} indicates a field to be filled by prompt data or metadata.

\begin{userinput}{MMLU Prompt Template:}
\texttt{The following are multiple choice questions (with answers) about \{\{domain\}\}. Think step by step and then finish your answer with "the answer is (X)" where X is the correct letter choice.}
\begin{verbatim}
Question: {{question}}
Options:
{{letter}}. {{choice}}
{{letter}}. {{choice}}
{{letter}}. {{choice}}
...
\end{verbatim}
\end{userinput}

\begin{userinput}{MATH Prompt Template:}
\begin{verbatim}
You are a highly skilled mathematician tasked with solving complex math problems.
Your goal is to provide clear, step-by-step solutions that can be easily parsed and
evaluated.

Here is the math problem you need to solve:

<problem>
{{MATH_PROBLEM}}
</problem>

Box your final answer using LaTeX, for example: $x = \\boxed{[Your final numerical or
algebraic answer]}$.

Now, please solve the given math problem and provide your solution in the specified format.
\end{verbatim}
\end{userinput}

\begin{userinput}{GPQA Prompt Template:}
\texttt{The following is a \{\{domain\}\} multiple choice question. Think step by step and then finish your answer with "the answer is (X)" where X is the correct letter choice.}
\begin{verbatim}
Question: {{question}}
Choices:
(A) {{choice1}}
(B) {{choice2}}
(C) {{choice3}}
(D) {{choice4}}
\end{verbatim}

\end{userinput}

\begin{userinput}{MBPP-Plus Prompt Template:}
\begin{verbatim}
Below will be an instruction to write a python function that accomplishes a task. 
You will also be given starter code with a function definition and any required imports. 
Think step-by-step, write comments clearly explaining each part of the code, and make sure 
your code solution is enclosed in markdown ticks (``` [your code here] ```).

<instruction>
{{instruction}}
</instruction>

<starter_code>
```
{{starter_code}}
    pass
```
</starter_code>
\end{verbatim}
\end{userinput}

\begin{userinput}{IFEval Prompt Template:}
\begin{verbatim}
It is extemely important that you follow all instructions exactly:
{{prompt}}
\end{verbatim}
\end{userinput}

\subsubsection{Detailed Scores}

\begin{table}[H]
\centering
\resizebox{\textwidth}{!}{
\begin{tabular}{lcccccc}
\toprule
Reward Model & MMLU Pro & Math & GPQA & MBPP Plus & IF Eval & Mean \\
\midrule
Athene-RM-70B & 0.761 & 0.607 & 0.499 & 0.748 & 0.633 & 0.650 \\
InternLM2-20B-Reward & 0.673 & 0.538 & 0.471 & 0.654 & 0.652 & 0.598 \\
Llama-3-Offsetbias-RM-8B & 0.590 & 0.481 & 0.450 & 0.819 & 0.646 & 0.597 \\
Athene-RM-8B & 0.656 & 0.517 & 0.459 & 0.675 & 0.586 & 0.579 \\
Nemotron-4-340B-Reward & 0.697 & 0.499 & 0.484 & 0.567 & 0.623 & 0.574 \\
InternLm2-7B-Reward & 0.638 & 0.552 & 0.457 & 0.562 & 0.658 & 0.573 \\
ArmoRM-Llama3-8B-v0.1 & 0.654 & 0.508 & 0.470 & 0.602 & 0.601 & 0.567 \\
Skywork-Reward-Llama-3.1-8B & 0.641 & 0.500 & 0.468 & 0.581 & 0.639 & 0.566 \\
Starling-RM-34B & 0.651 & 0.476 & 0.453 & 0.634 & 0.569 & 0.557 \\
Eurus-RM-7B & 0.607 & 0.516 & 0.438 & 0.590 & 0.594 & 0.549 \\
Skywork-Reward-Gemma-2-27B & 0.550 & 0.462 & 0.447 & 0.691 & 0.583 & 0.547 \\
InternLM2-1-8B-Reward & 0.538 & 0.411 & 0.451 & 0.572 & 0.581 & 0.510 \\
Starling-RM-7B-Alpha & 0.562 & 0.409 & 0.433 & 0.559 & 0.564 & 0.505 \\
NaiveVerbosityModel & 0.487 & 0.349 & 0.420 & 0.568 & 0.539 & 0.473 \\
\bottomrule
\end{tabular}
}
\caption{Reward Model Best of K Performance Across Benchmarks}
\vspace{-0.5em}
\label{tab:rm_bok_performance}
\end{table}

\begin{table}[H]
\centering
\resizebox{\textwidth}{!}{
\begin{tabular}{lcccccc}
\toprule
Reward Model & MMLU Pro & Math & GPQA & MBPP Plus & IF Eval & Mean \\
\midrule
Athene-RM-70B & 0.792 & 0.760 & 0.603 & 0.661 & 0.594 & 0.682 \\
Internlm2-20B-reward & 0.677 & 0.691 & 0.562 & 0.574 & 0.595 & 0.620 \\
Llama-3-offsetbias-RM-8B & 0.631 & 0.617 & 0.541 & 0.710 & 0.594 & 0.619 \\
Athene-RM-8B & 0.683 & 0.673 & 0.560 & 0.602 & 0.556 & 0.615 \\
Nemotron-4-340B-Reward & 0.704 & 0.660 & 0.570 & 0.506 & 0.587 & 0.605 \\
Skywork-Reward-Llama-3.1-8B & 0.663 & 0.678 & 0.560 & 0.523 & 0.586 & 0.602 \\
Internlm2-7B-Reward & 0.665 & 0.718 & 0.558 & 0.464 & 0.605 & 0.602 \\
ArmoRM-Llama3-8B-v0.1 & 0.678 & 0.659 & 0.549 & 0.538 & 0.573 & 0.599 \\
Starling-RM-34B & 0.683 & 0.621 & 0.547 & 0.534 & 0.536 & 0.584 \\
Eurus-RM-7B & 0.627 & 0.665 & 0.521 & 0.537 & 0.554 & 0.581 \\
Skywork-Reward-Gemma-2-27B & 0.542 & 0.582 & 0.506 & 0.572 & 0.536 & 0.547 \\
Internlm2-1-8B-Reward & 0.561 & 0.587 & 0.538 & 0.462 & 0.538 & 0.537 \\
Starling-RM-7B-Alpha & 0.547 & 0.527 & 0.506 & 0.400 & 0.519 & 0.500 \\
NaiveVerbosityModel & 0.495 & 0.528 & 0.506 & 0.330 & 0.511 & 0.474 \\
\bottomrule
\end{tabular}
}
\caption{Area Under ROC Curve for Reward Models across Benchmarks}
\vspace{-0.5em}
\label{tab:area_roc_curve}
\end{table}

\begin{table}[h!]
\centering
\setlength{\tabcolsep}{3pt}
\resizebox{\textwidth}{!}{
\begin{tabular}{lcccccccccccc}
\toprule
\multirow{2}{*}[-0.5em]{{Reward Model}} & 
\multicolumn{3}{c}{{gemma-2-9b-it}} & 
\multicolumn{3}{c}{{gpt-4o-mini}} & 
\multicolumn{3}{c}{Llama-3-8B} & 
\multicolumn{3}{c}{claude-3-haiku} \\
\cmidrule(lr){2-4} \cmidrule(lr){5-7} \cmidrule(lr){8-10} \cmidrule(lr){11-13}
 & Loss & Max & End & Loss & Max & End & Loss & Max & End & Loss & Max & End \\
\midrule
athene-rm-70b & 0.093 & 0.702 & 0.681 & 0.110 & 0.678 & 0.629 & 0.113 & 0.669 & 0.653 & 0.131 & 0.633 & 0.605 \\
armorm-llama3-8b-v0.1 & 0.119 & 0.657 & 0.636 & 0.147 & 0.620 & 0.580 & 0.179 & 0.576 & 0.537 & 0.194 & 0.564 & 0.512 \\

naiveverbositymodel & 0.241 & 0.508 & 0.463 & 0.250 & 0.554 & 0.425 & 0.358 & 0.448 & 0.317 & 0.337 & 0.467 & 0.355 \\
eurus-rm-7b & 0.143 & 0.627 & 0.597 & 0.158 & 0.613 & 0.562 & 0.187 & 0.562 & 0.512 & 0.228 & 0.531 & 0.452 \\
skywork-reward-gemma-2-27b & 0.169 & 0.583 & 0.543 & 0.175 & 0.590 & 0.549 & 0.209 & 0.534 & 0.494 & 0.190 & 0.558 & 0.529 \\
skywork-reward-llama-3.1-8b & 0.126 & 0.643 & 0.612 & 0.136 & 0.633 & 0.597 & 0.189 & 0.565 & 0.527 & 0.216 & 0.561 & 0.491 \\
llama-3-offsetbias-rm-8b & 0.133 & 0.653 & 0.629 & 0.146 & 0.629 & 0.585 & 0.210 & 0.542 & 0.502 & 0.151 & 0.620 & 0.592 \\
nemotron-4-340b-reward & 0.129 & 0.641 & 0.617 & 0.128 & 0.644 & 0.618 & 0.159 & 0.610 & 0.583 & 0.232 & 0.565 & 0.485 \\
starling-rm-34b & 0.157 & 0.602 & 0.570 & 0.151 & 0.622 & 0.563 & 0.183 & 0.562 & 0.528 & 0.209 & 0.545 & 0.487 \\
athene-rm-8b & 0.142 & 0.621 & 0.584 & 0.133 & 0.636 & 0.600 & 0.175 & 0.589 & 0.543 & 0.183 & 0.560 & 0.531 \\

internlm2-7b-reward & 0.138 & 0.630 & 0.588 & 0.147 & 0.633 & 0.581 & 0.155 & 0.608 & 0.581 & 0.253 & 0.565 & 0.462 \\
starling-rm-7b-alpha & 0.183 & 0.569 & 0.535 & 0.199 & 0.578 & 0.516 & 0.238 & 0.508 & 0.476 & 0.319 & 0.486 & 0.378 \\
internlm2-1-8b-reward & 0.193 & 0.566 & 0.501 & 0.191 & 0.583 & 0.506 & 0.218 & 0.526 & 0.480 & 0.256 & 0.503 & 0.448 \\
internlm2-20b-reward & 0.124 & 0.648 & 0.626 & 0.130 & 0.646 & 0.607 & 0.159 & 0.602 & 0.570 & 0.166 & 0.586 & 0.570 \\
\bottomrule
\end{tabular}
}
\caption{Average Best of K per Sample Model across MMLU Pro, Math, GPQA, MBPP Plus, and IF Eval}
\vspace{-0.5em}
\label{tab:avg_bok_per_sample}

\end{table}

\begin{table}[H]
\centering
\resizebox{\textwidth}{!}{
\begin{tabular}{lcccccc}
\toprule
Reward Model & gemma-2-9b-it & gpt-4o-mini & Llama-3-8B & claude-3-haiku \\
\midrule
athene-rm-70b & 0.710 & 0.648 & 0.710 & 0.674 \\
armorm-llama3-8b-v0.1 & 0.655 & 0.577 & 0.616 & 0.591 \\
naiveverbositymodel & 0.515 & 0.491 & 0.487 & 0.433 \\
eurus-rm-7b & 0.620 & 0.546 & 0.621 & 0.562 \\
skywork-reward-gemma-2-27b & 0.553 & 0.519 & 0.562 & 0.550 \\
skywork-reward-llama-3.1-8b & 0.639 & 0.594 & 0.619 & 0.578 \\
llama-3-offsetbias-rm-8b & 0.628 & 0.574 & 0.583 & 0.650 \\
nemotron-4-340b-reward & 0.639 & 0.586 & 0.658 & 0.561 \\
starling-rm-34b & 0.602 & 0.571 & 0.604 & 0.574 \\
athene-rm-8b & 0.640 & 0.592 & 0.635 & 0.601 \\

internlm2-7b-reward & 0.657 & 0.573 & 0.655 & 0.569 \\
starling-rm-7b-alpha & 0.544 & 0.499 & 0.525 & 0.475 \\
internlm2-1-8b-reward & 0.581 & 0.536 & 0.570 & 0.504 \\
internlm2-20b-reward & 0.629 & 0.603 & 0.650 & 0.603 \\
\bottomrule
\end{tabular}
}
\caption{Average AUC per sample model across MMLU Pro, Math, GPQA, MBPP Plus, and IF Eval}
\vspace{-0.5em}
\label{tab:avg_auc_table}
\end{table}

\begin{figure}[H]
    \centering
    \includegraphics[width=1.0\linewidth]{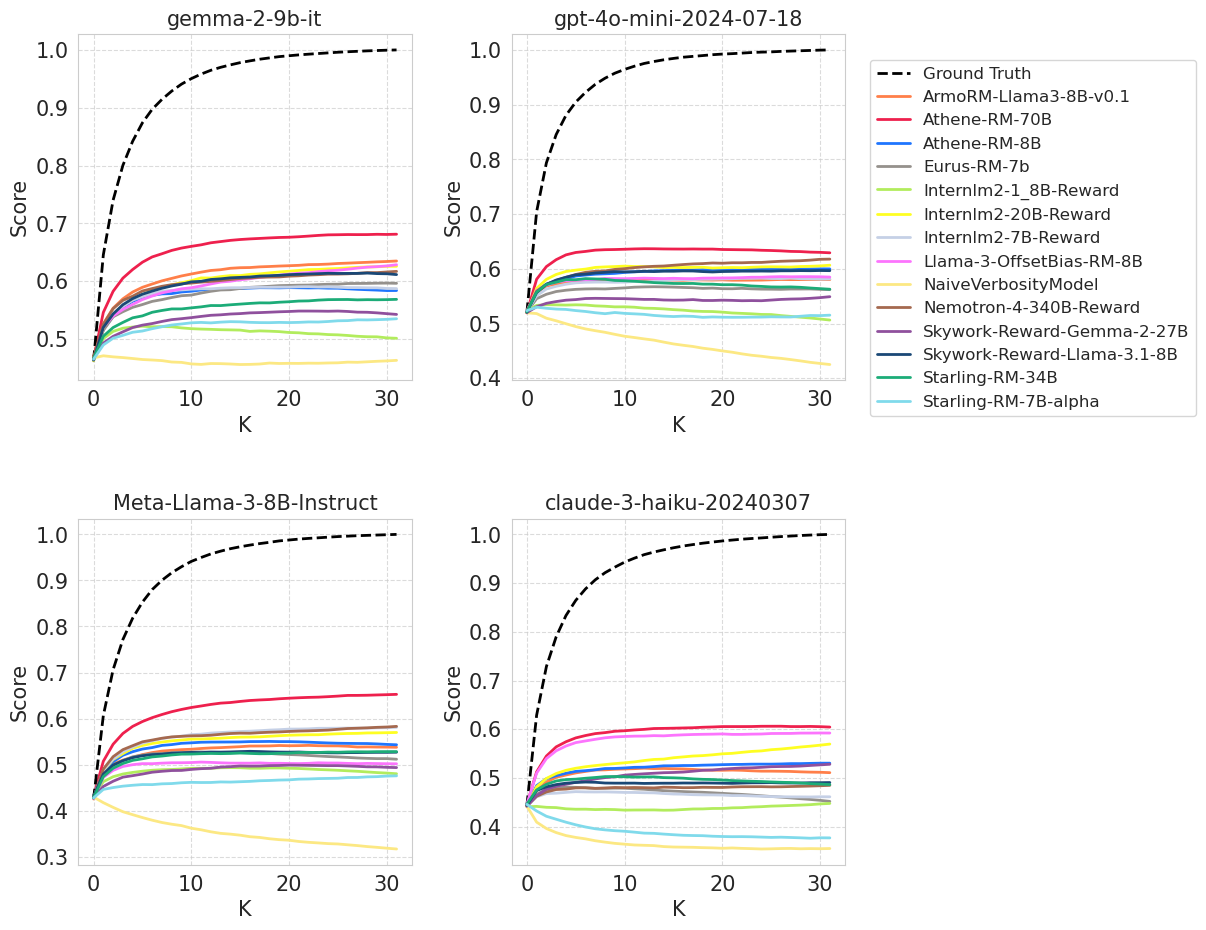}
    \caption{Performance average across all benchmarks, conditioned on each sample model}
    \label{fig:sample model performance graph}
\end{figure}

\begin{figure}[H]
    \centering
    \includegraphics[width=1\linewidth]{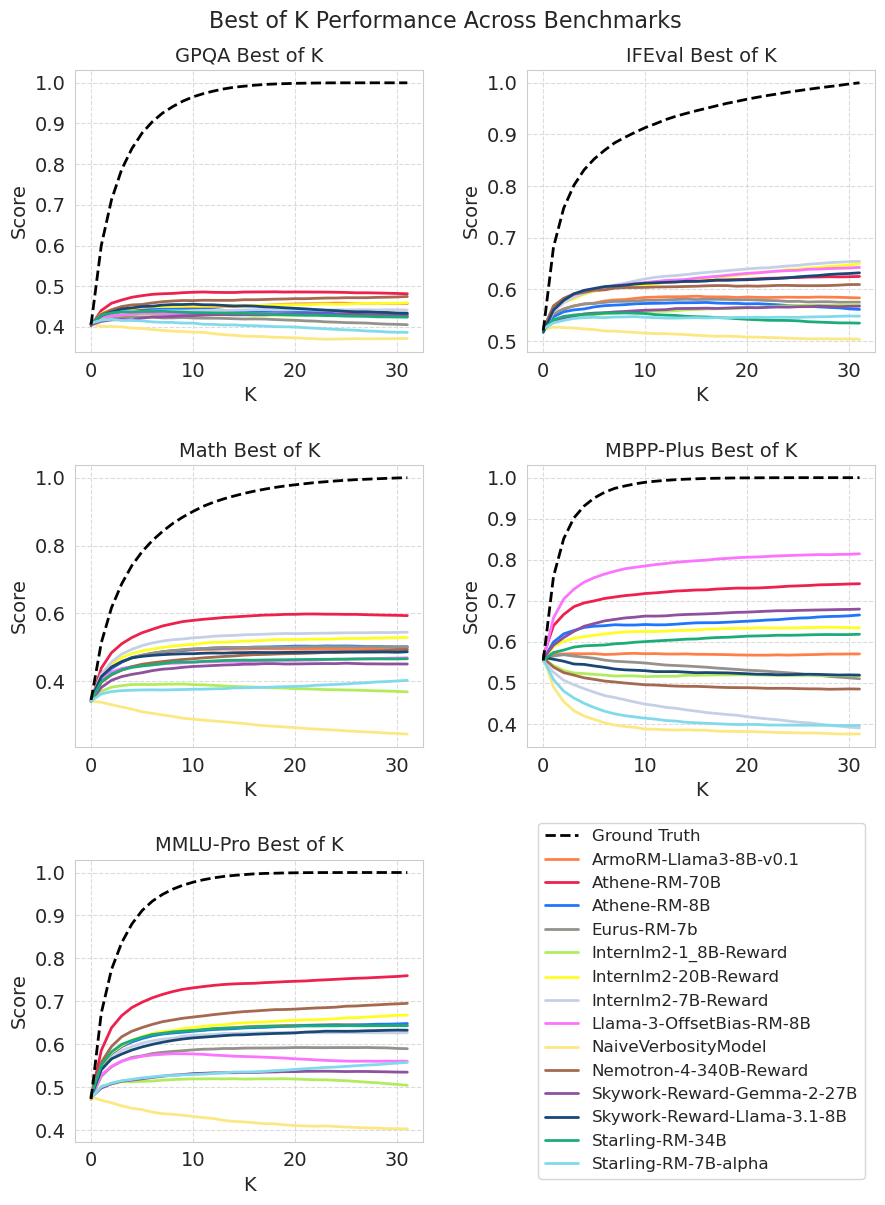}
    \caption{Performance comparison across all benchmarks}
    \label{fig:benchmark graph comparisons}
\end{figure}

\begin{figure}[H]
    \centering
    \includegraphics[width=1\linewidth]{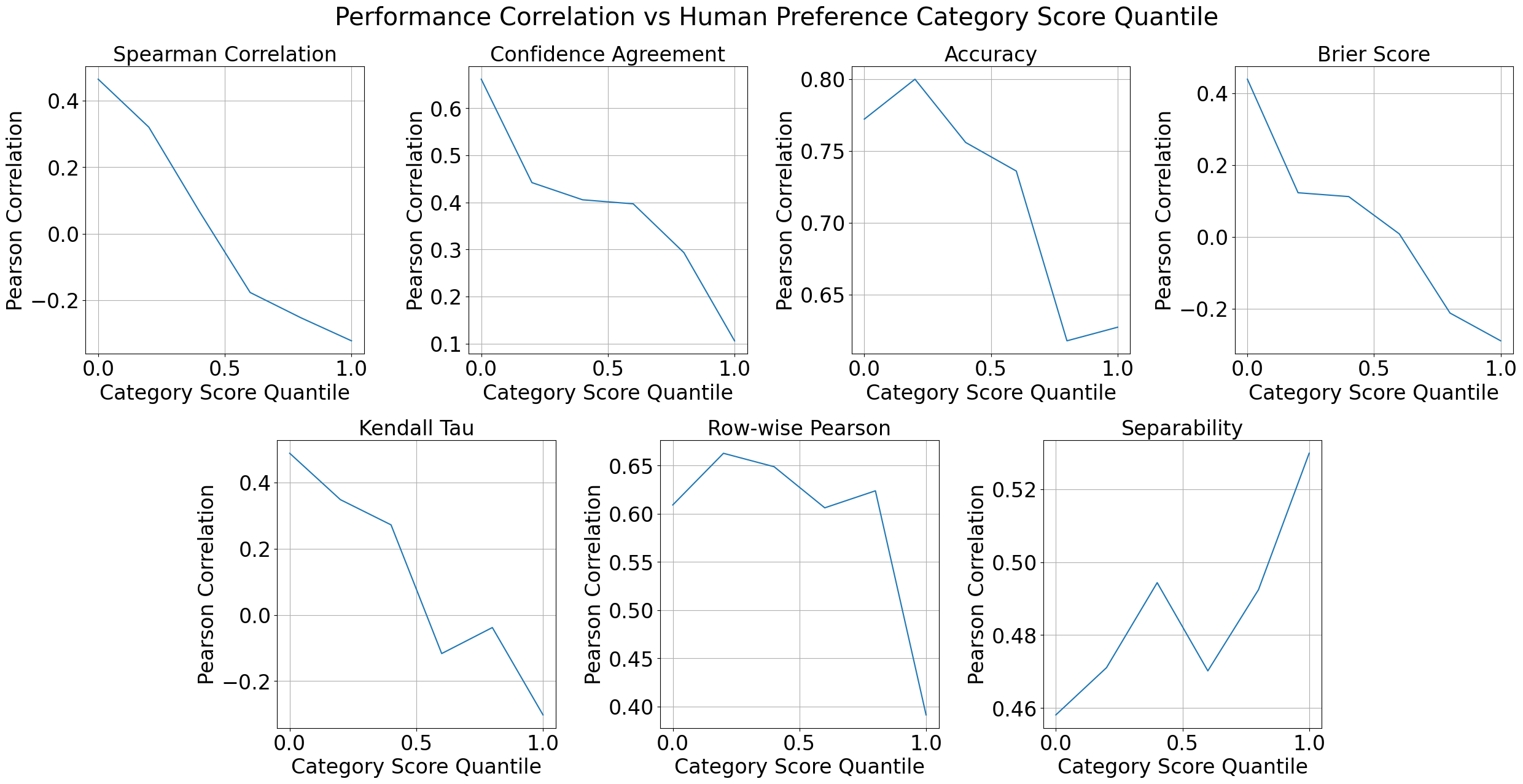}
    \caption{The graphs show all metrics for the human preference dataset. For each metric, the six benchmarks (Hard, Easy, Instruction Following, Coding, Math, and Similar Responses Prompts) (all mean and SD normalized) aggregated into final score by quantile (x-axis). The Pearson Correlation between the aggregated scores are calculated relative to Post-RLHF Human Preference Arena Score ratings for each aggregation level. Notice that for all metrics except Separability, decreasing quantile increases correlation.
    }
    \label{fig:full aggregation plots}
\end{figure}
\subsection{DPO Configuration}

\begin{table}[H]
\begin{tabular}{r|l}
DPO Configuration  &                               \\ \hline
Base Model         & Meta-Llama-3.1-8B-Instruct    \\
$\tau$                & 0.1                           \\
Learning Rate      & $2.00\times10^{-0.6}$                      \\
LR Schedule        & Constant                      \\
Global Batch Size  & 64                            \\
Max Length         & 8192                          \\
Max Prompt Length  & 4096                          \\
Implementation     & TRL DPOTrainer \citep{vonwerra2022trl}                \\
Optimizer          & AdamW, $\beta_1=0.9$, $\beta_2=0.999$ \\
Space Optimization & Deepspeed Zero2              
\end{tabular}
\end{table}

\subsection{Crowdsourced Human Preference Vote Details} \label{appendix: C}
\begin{table}[H]
\resizebox{\textwidth}{!}{
\begin{tabular}{l|l|l|l|l|l}
\#Votes                    & Est. Unique Users         & Mean Votes/User           & Median Votes/User      & Mean Battles/Pair           & Mean Votes/Model            \\ \hline
\multicolumn{1}{r|}{12190} & \multicolumn{1}{r|}{6120} & \multicolumn{1}{r|}{1.99} & \multicolumn{1}{r|}{1.00} & \multicolumn{1}{r|}{190.47} & \multicolumn{1}{r}{2031.67}
\end{tabular}}
\caption{Statistics on vote participation and distribution for crowdsourced human preference labels.}
\label{tab:human-pref-votes}
\end{table}

\subsection{Additional Analysis on Downstream Performance} \label{appd-style-control}

\begin{table}[H]
\centering
\resizebox{0.8\textwidth}{!}{
\begin{tabular}{lrrr}
\toprule
Model & Arena Score & 95\% CI Lower & 95\% CI Upper \\
\midrule
Meta-Llama-3.1-70B-Instruct\textsuperscript{*} & 1229 & 1218 & 1239 \\
Athene-RM-70B & \textbf{1209} & 1201 & 1218 \\
Athene-RM-8B & 1203 & 1194 & 1211 \\
internlm2-7b-reward & 1201 & 1192 & 1210 \\
Llama-3-OffsetBias-RM-8B & 1197 & 1188 & 1204 \\
ArmoRM-Llama3-8B-v0.1 & 1185 & 1175 & 1191 \\
Meta-Llama-3.1-8B-Instruct\textsuperscript{*} & 1177 & 1168 & 1186 \\
Skywork-Reward-Llama-3.1-8B & 1171 & 1163 & 1182 \\
Nemotron-4-340B-Reward & 1170 & 1161 & 1180 \\
internlm2-20b-reward & 1170 & 1159 & 1179 \\
Skywork-Reward-Gemma-2-27B & 1170 & 1160 & 1180 \\
Meta-Llama-3-8B-Instruct\textsuperscript{*} & 1152 & 1142 & 1160 \\
\bottomrule
\end{tabular}}
\caption{Post DPO performance on Chatbot Arena Overall Category after applying style-control. “Model” is the reward model
used to train the base model. Models marked with “*” are baseline unaltered models. The best
non-base model Arena Score is bolded.}
\label{tab:style-control-elo}
\end{table}

\begin{figure}[H]
    \centering
    \includegraphics[width=1\linewidth]{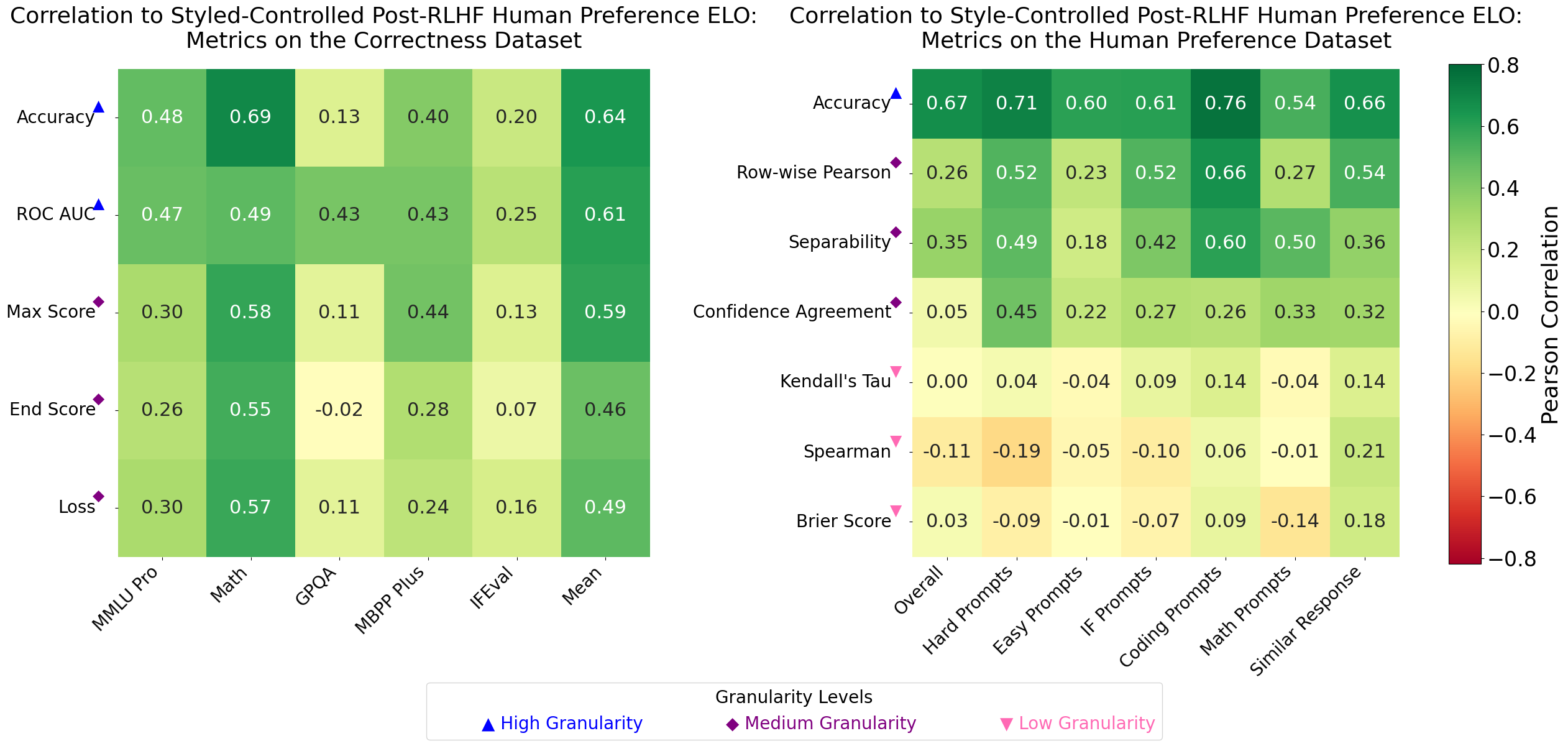}
    \caption{Pearson correlations between various metrics and styled-controlled human preference scores. Left: Correlations between metrics on the Correctness Dataset and Post-RLHF human preference Arena Score. Right: Correlations between metrics on the Human Preference Dataset and Post-RLHF human preference Arena Score.}
    \label{fig:sccorrs-cap}
\end{figure}

\begin{figure}[H]
    \centering
    \includegraphics[width=0.4\linewidth]{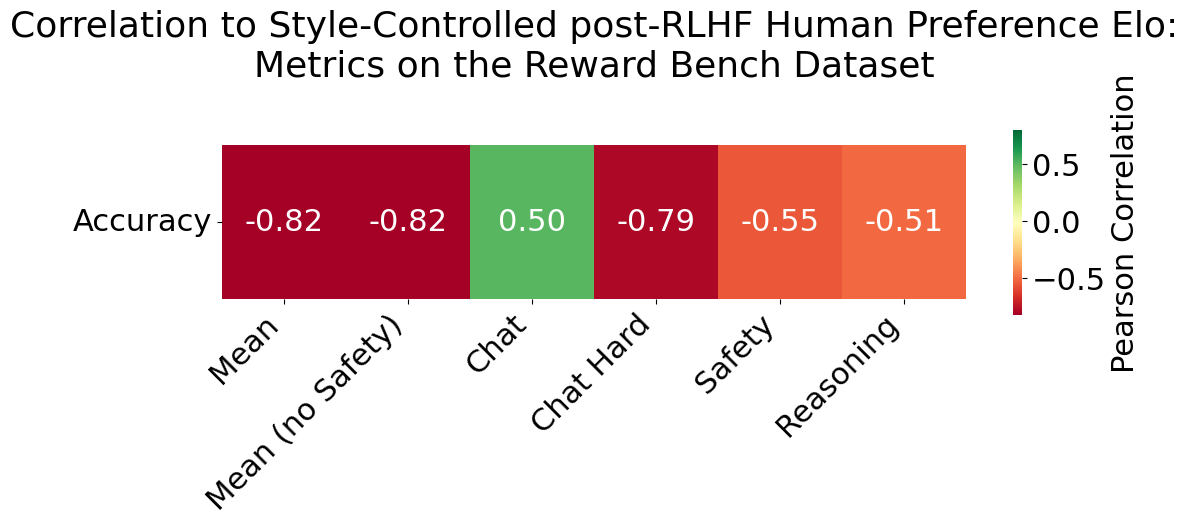}
    \caption{Pearson correlation between the ranking of models in RewardBench and their respective style-controlled Post-DPO Arena Score rankings on real human preference.}
    \label{fig:screward-bench-correlations}
\end{figure}

As an ablation, we calculate style-controlled human preference Arena Scores. Style-controlled Arena Scores fit the Bradley Terry model with style elements as features of the regression. These features are used to decouple style from model Arena Scores; this process yields score estimates, style 
\textit{aside}. The full process for style control is detailed in \citet{stylearena2024}. For maximum coverage, we control for length and markdown.
\end{document}